\documentclass[10pt,twocolumn,letterpaper]{article}

\usepackage[pagenumbers]{cvpr} 

\definecolor{cvprblue}{rgb}{0.21,0.49,0.74}
\usepackage[pagebackref,breaklinks,colorlinks,allcolors=cvprblue]{hyperref}
\usepackage{array}
\usepackage[table]{xcolor}
\usepackage[ruled,vlined]{algorithm2e}
\usepackage{algorithm2e}
\usepackage{pifont} 
\usepackage{xcolor}
\usepackage{listings}
\def\paperID{6666666666} 
\def\confName{CVPR}
\def\confYear{2026}

\title{Spatial Policy: Guiding Visuomotor Robotic Manipulation\\
with Spatial-Aware Modeling and Reasoning}

\author{
    Yijun Liu\textsuperscript{1},
    Yuwei Liu\textsuperscript{2},
    Yuan Meng\textsuperscript{1},
    Jieheng Zhang\textsuperscript{3},
    Yuwei Zhou\textsuperscript{1},\\
    Ye Li\textsuperscript{1},
    Jiacheng Jiang\textsuperscript{1},
    Kangye Ji\textsuperscript{1},
    Shijia Ge\textsuperscript{1},
    Zhi Wang\textsuperscript{1},
    Wenwu Zhu\textsuperscript{1}\\[4pt]
    \textsuperscript{1}Tsinghua University \quad
    \textsuperscript{2}Beijing University of Technology \quad
    \textsuperscript{3}Guangzhou University
}

\begin{document}
\maketitle
\begin{abstract}
Vision-centric hierarchical embodied models have demonstrated strong potential.
However, existing methods lack spatial awareness capabilities, limiting their effectiveness in bridging visual plans to actionable control in complex environments.
To address this problem,
we propose Spatial Policy
(SP), a unified spatial-aware visuomotor robotic manipulation framework via explicit spatial modeling and reasoning.
Specifically, we first design a spatial-conditioned embodied video generation module to model spatially guided predictions through the spatial plan table.
Then, we propose a flow-based action prediction module to infer executable actions with coordination.
Finally, we propose a spatial reasoning feedback policy to refine the spatial plan table via dual-stage replanning.
Extensive experiments show that SP substantially outperforms state-of-the-art baselines, achieving over 33\% improvement on Meta-World and over 25\% improvement on iTHOR, demonstrating strong effectiveness across 23 embodied control tasks. We additionally evaluate SP in real-world robotic experiments to verify its practical viability.
SP enhances the practicality of embodied models for robotic control applications.
Code and checkpoints are maintained at https://plantpotatoonmoon.github.io/SpatialPolicy/.
\end{abstract}    
\section{Introduction}
In recent years, hierarchical embodied models combining high-level planning and low-level action have achieved remarkable success in robotic manipulation~\cite{rt2,gr2,dp3d,spvla}.
Among these approaches, vision-based methods, known as visuomotor robotic manipulation, have emerged as a particularly promising direction. By leveraging rich visual cues, these methods enable effective cross-hierarchical information transfer across diverse robotic embodiments and environments~\cite{unipi,seer,avdc,videoagent,susie,grmg}.
Specifically, visuomotor robotic manipulation employs a generative model to predict trajectories of future video, which are subsequently translated into executable actions through an action prediction module.

Despite significant advances, existing visuomotor robotic manipulation frameworks lack spatial awareness abilities, failing to effectively bridge visual plans to executable actions in complex environments~\cite{seer,susie}. 
For instance, AVDC~\cite{avdc} lacks explicit spatial modeling in its video generation process, resulting in physically implausible predictions, such as robotic arms passing through walls.
Feedback module of CLOVER~\cite{clover} enhances dynamic environment adaptation, but its inability to process spatial relationships prevents handling of occluded target scenarios~\cite{seer}.

To tackle this problem, we investigate the spatial-aware visuomotor robotic manipulation via explicit spatial modeling and reasoning,
which introduces three key challenges: 
(1) how to generate videos that explicitly model spatial relations, such as object-goal positioning and task alignment; (2) how to utilize the generated videos to guide action prediction in a way that preserves spatial consistency and task relevance; and (3) how to reason over spatial cues during execution to enable real-time feedback that effectively bridges visual and control.

To address these challenges,
we propose Spatial Policy (\textbf{SP}), a unified framework for robotic manipulation that integrates spatial-aware video generation, action prediction, and feedback.
The core idea is to unify spatial modeling and reasoning via a spatial plan table that guides video generation, enables spatially consistent action prediction, and supports feedback-driven refinement during execution.

Specifically,
(1) we propose a Spatial-Conditioned Embodied Video Generation module that generates visual predictions guided by a structured spatial plan table, which encodes object-goal relations and directional cues to ensure spatial coherence in imagined futures, addressing the challenge of modeling spatially meaningful video trajectories;
(2) we introduce a Flow-Based Action Prediction module that infers executable actions by capturing spatial motion dynamics between predicted frames using flow and spatial coordinates, tackling the challenge of translating visual predictions into spatially consistent actions;
(3) we develop a Spatial Reasoning Feedback Policy that refines the spatial plan and video execution through dual-stage replanning based on VLM feedback and rule-based corrections, solving the challenge of real-time spatial reasoning during policy execution.

Extensive experiments show that our model surpasses existing baselines, achieving 86.7\% performance on Meta-World and 59.6\% on iTHOR and further proving its viability through real-world robotic experiments. The contributions of this paper are summarized as follows:
\begin{itemize}
    \item \textbf{Spatial policy for vision-action alignment.} We propose SP, a spatially grounded visuomotor policy framework that explicitly models and reasons about spatial structures to align visual prediction with executable actions.

    \item \textbf{Modular design with structured spatial modeling and reasoning.} SP consists of three synergistic modules: a spatial-conditioned video generation module guided by a structured spatial plan table, a flow-based action prediction that preserves spatial consistency, and a spatial reasoning feedback policy that refines execution through real-time spatial replanning.

    \item \textbf{Comprehensive evaluation and strong performance.} We conduct extensive experiments on the MetaWorld benchmark to assess the impact of spatial integration, including ablations on video generation, action prediction, and feedback mechanisms. SP yields an absolute improvement of over 33\% on Meta-World and over 25\% on iTHOR and futher validate our model on real-world robotic executions, demonstrating the effectiveness of spatial reasoning in visuomotor robotic manipulation.
\end{itemize}

\section{Related Work}
\subsection{Video Generation for Embodied Manipulation}

Video generation has recently been explored as a visual planning mechanism for embodied agents. AVDC~\cite{avdc} generates action-annotated video plans from raw RGB inputs using dense correspondencesenabling policy learning without action supervision. However, due to the lack of explicit spatial reasoning in both generation and action prediction, limiting precision in manipulation tasks. VideoAgent~\cite{videoagent} refines video plans via self-conditioning to reduce hallucinations and improve visual quality. However, it inherits the lack of explicit spatial reasoning from AVDC in both generation and action prediction, limiting precision in geometry-sensitive manipulation tasks. Clover~\cite{clover} introduces a closed-loop framework combining video diffusion planning and feedback control via an embedding-based error signal, achieving strong results in long-horizon tasks. However, its spatial modeling is limited to depth-based cues and lacks explicit 3D spatial planning or structured geometric reasoning, which restricts generalization across diverse spatial layout. SEER~\cite{seer} proposes a Predictive Inverse Dynamics Model (PIDM) that forecasts future visual states and infers actions end-to-end, closing the loop between vision and control. While it achieves strong performance through large-scale training, it lacks explicit spatial reasoning modules and does not leverage structured spatial plans, which may limit interpretability and adaptability in geometry-sensitive tasks. SuSIE~\cite{susie} leverages an image-editing diffusion model to generate visual subgoals for a low-level goal-conditioned policy, enabling better generalization in unstructured environments. However, its subgoal generation relies on appearance-based edits without explicit incorporation of 3D spatial relationships, limiting its precision in spatially complex manipulation scenarios.

\subsection{Spatial Reasoning in Vision-Language Embodied Models}

Spatial understanding has been addressed in embodied AI from several perspectives. SpatialVLM~\cite{spatialvlm} and GraspCorrect~\cite{graspcorrect} leverage pretrained vision-language models to extract chain-of-thought spatial cues and determine object relations in manipulation tasks. 3D-VLA~\cite{3dvla} explicitly encodes spatial semantics by aligning 3D vision with textual instructions. SPA~\cite{spa} utilizes point cloud reconstruction for spatial alignment in manipulation and navigation tasks. Meanwhile, LLM-Planner  \cite{llmplanner} and related methods employ large language models to infer spatially structured plans in symbolic form.
Despite these advances, existing methods primarily address semantic or symbolic spatial reasoning and lack visual feedback coupled with executable rollouts. Many rely on handcrafted prompts or fixed spatial priors and do not support continuous video prediction. Our approach fills this gap by embedding structured spatial plans, automatically derived from the environment geometry, into a generative video prediction framework, thereby integrating spatial reasoning and visual synthesis for action-aware planning.



\begin{figure*}[t]
  \centering
  \includegraphics[width=\textwidth]{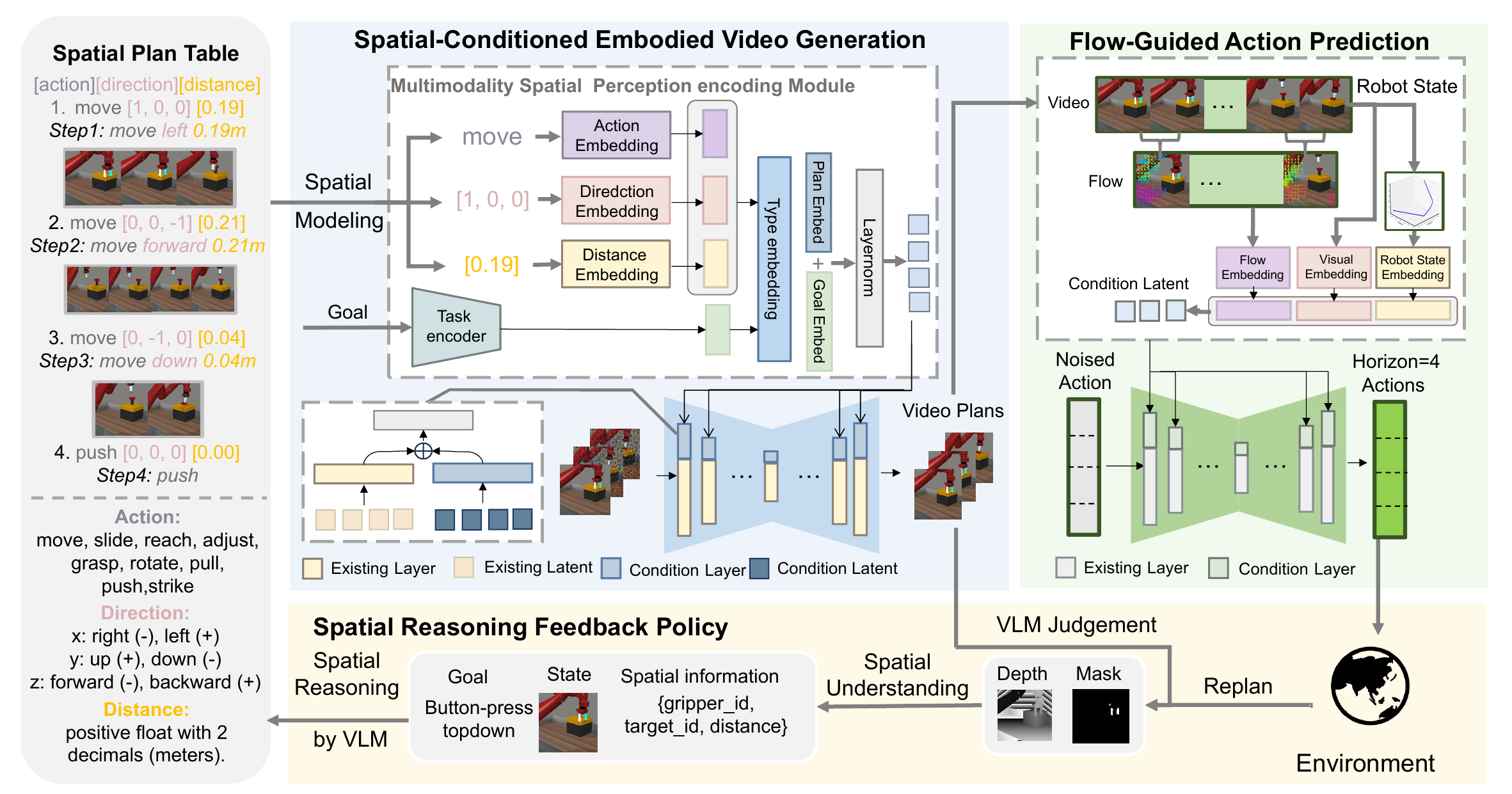} 
  \caption{\textbf{Framework Overview of Spatial Policy.} Our system comprises three modules: 
(1) \textbf{Spatial-Conditioned Embodied Video Generation}, which uses a \textbf{Spatial Plan Table} obtained via VLM reasoning over the spatial offset between the robot and the object, to guide diffusion model in generating spatially coherent video prediction.
(2) \textbf{Flow-Based Action Prediction}, which converts the generated video into executable actions using flow and spatial coordinates. 
(3) \textbf{Spatial Reasoning Feedback Policy}, which enables real-time correction via a dual-stage replanning strategy, combining VLM-based video judgement and policy diagnostics to refine the spatial plan table for closed-loop control.}

  \label{fig:framework}
\end{figure*}

\section{Spatial Policy}
We propose a visuomotor robotic manipulation framework that integrates structured spatial reasoning into the perception-to-control pipeline, enabling robots to generate spatially consistent visual plans and execute them reliably. As illustrated in Fig.~\ref{fig:framework}, our system comprises three interconnected modules.
First, the \textbf{Spatial-Conditioned Embodied Video Generation} module in Sec.~\ref{3.1} generates future video trajectories conditioned on a structured spatial plan table composed of atomic actions, directional vectors, and relative distances.
Second, the \textbf{Flow-Based Action Prediction} module in Sec.~\ref{3.2} predicts actions aligned with the generated video using the flow and spatial coordinates.
Third, the \textbf{Spatial Reasoning Feedback Policy} monitors spatial alignment during execution and dynamically refines both the spatial plan table and the visual predictions.
Together, these modules form a closed-loop framework that maintains spatial consistency across video imagination, action prediction, and feedback-based reasoning. The full inference pipeline is provided in Appendix A.1.

\subsection{Spatial-Conditioned Embodied Video Generation}
\label{3.1}
To generate spatially grounded visual trajectories for visuomotor robotic manipulation, we adopt a conditional diffusion framework for future video prediction. The model takes as input an initial observation frame and a textual task description, and generates a sequence of future frames representing high-level visual plans. To enhance spatial awareness, we additionally introduce the spatial plan table derived from spatial relationships as an auxiliary global condition.

\paragraph{Spatial Plan Table}

To capture precise spatial intent, we extract the coordinates of the robot end-effector $\mathbf{p}_{\text{ee}} \in \mathbb{R}^3$ and the target object $\mathbf{p}_{\text{obj}} \in \mathbb{R}^3$ from the simulator, and compute their relative offset as $\Delta \mathbf{p} = \mathbf{p}_{\text{obj}} - \mathbf{p}_{\text{ee}}$. This spatial offset is passed into a pretrained visual-language model (GPT-4o) to produce a structured spatial plan. The generated plan consists of sequential subgoals, each encoding an atomic action with an action type, a direction vector, and a distance scalar, as illustrated in Fig.~\ref{fig:framework}. The detailed VLM inference procedure is provided in Appendix A.5.

\paragraph{Encoding and Integration}

Each subgoal is independently encoded into a fixed-dimensional latent vector. Concretely, the action type is mapped to an embedding $\mathbf{e}_{\text{act}} \in \mathbb{R}^d$ via a learnable embedding table, the direction vector is discretized and embedded into $\mathbf{e}_{\text{dir}} \in \mathbb{R}^d$ using another learnable embedding table, and the distance scalar is projected into $\mathbf{e}_{\text{dis}} \in \mathbb{R}^d$ through a learnable MLP. The combined subgoal embedding is given by:
\begin{equation}
\mathbf{e}_{\text{subplan}} = \text{MLP}\left([\mathbf{e}_{\text{act}};\, \mathbf{e}_{\text{dir}};\, \mathbf{e}_{\text{dis}}] \right),
\label{eq:subplan-embed}
\end{equation}

All subgoal embeddings $\{\mathbf{e}_1, \dots, \mathbf{e}_n\}$ are concatenated with the task text embedding $\mathbf{z}_{\text{task}} \in \mathbb{R}^d$ obtained via CLIP~\cite{clip} to form the composite global condition:
\begin{equation}
\mathcal{T} = \left[ \mathbf{z}_{\text{task}};\, \mathbf{e}_1;\, \dots;\, \mathbf{e}_n \right].
\label{eq:global-condition}
\end{equation}

This global condition is injected into all layers of the U-Net through conditional normalization mechanisms, guiding the diffusion model to produce spatially grounded, instruction-aligned video sequences.

\paragraph{Conditional Diffusion Formulation}

We follow the design of a U-Net-based denoising diffusion probabilistic model, adapted from the Imagen architecture~\cite{imagen}, consistent with AVDC~\cite{avdc}. Given an initial frame $\mathbf{I}_0 \in \mathbb{R}^{H \times W \times 3}$ and a composite condition vector $\mathcal{T}$ that incorporates both task instructions and structured spatial information, the model predicts a sequence of $T=7$ future frames $\mathbf{I}_{1:T}$, resulting in an 8-frame trajectory. The generative objective is to approximate the posterior distribution:
\begin{equation}
p(\mathbf{I}_{1:T} \mid \mathbf{I}_0, \mathcal{T}),
\label{eq:video-gen-posterior}
\end{equation}
where the training target is to denoise a Gaussian-perturbed version of the future frames. Specifically, the model minimizes the following mean squared error (MSE) loss:
\begin{equation}
\mathcal{L}_{\text{MSE}} = \mathbb{E}_{\epsilon, t} \left\| \epsilon - \epsilon_\theta\left( \sqrt{1 - \beta_t} \cdot \mathbf{I}_{1:T} + \sqrt{\beta_t} \cdot \epsilon,\, t \mid \mathbf{I}_0, \mathcal{T} \right) \right\|^2,
\label{eq:video-gen-loss}
\end{equation}
where $\epsilon \sim \mathcal{N}(0, \mathbf{I})$ is sampled noise, $\beta_t$ is the diffusion schedule, and $\epsilon_\theta$ is the model's denoising network.

\subsection{Flow-Based Action Prediction}
\label{3.2}
To incorporate explicit motion cues into action generation, we extend the original module by adding flow as an additional conditioning signal, while preserving the overall framework and notation style.

\paragraph{Multimodal Spatial Input Encoding}

We condition our policy on four types of information: (1) the current image $\mathbf{I}_{\text{cur}}$, (2) the goal image $\mathbf{I}_{\text{goal}}$ (sampled from the generated visual plan within the next $k$ steps), (3) the spatial coordinate of the robotic arm $\mathbf{p}_{\text{cur}} \in \mathbb{R}^3$, and (4) the dense flow field computed between consecutive frames. The flow field is computed using the classical Farnebäck method~\cite{farneback}:
\begin{equation}
\mathbf{F}_{t} = \text{Flow}\!\left(\mathbf{I}_{t-1}, \mathbf{I}_{t}\right),
\end{equation}
and then encoded using a CNN-based encoder. All inputs are encoded into compact feature vectors through their respective encoders (ResNet for images and an MLP for spatial coordinates). The modality-specific features are subsequently concatenated to form the global condition vector used by the diffusion policy.

\paragraph{Diffusion Policy-based Action Prediction Module}

We formulate action prediction as a conditional generative process. Specifically, we adopt a U-net-based diffusion policy architecture to model the conditional distribution over action trajectories:
\begin{equation}
p_\theta(\mathbf{a}_{0:H-1} \mid \mathbf{I}_{\text{cur}}, \mathbf{I}_{\text{goal}}, \mathbf{p}_{\text{cur}}, \mathbf{F}_{t})
\end{equation}
where $\mathbf{a}_{0:H-1}$ denotes a sequence of $H$ actions, each including a 3-DoF end-effector pose and gripper control. We train the policy via denoising score matching, minimizing:
\begin{equation}
\mathcal{L}_{\text{MSE}} = 
\mathbb{E}_{\mathbf{a}, \epsilon, t}
\left\|
\epsilon - \epsilon_\theta\!\left(
\sqrt{1 - \beta_t} \cdot \mathbf{a}
+ \sqrt{\beta_t} \cdot \epsilon,
t \mid \mathbf{c}
\right)
\right\|^2
\end{equation}
where $\mathbf{c}$ denotes the concatenated global condition vector including all encoded modalities, $\epsilon \sim \mathcal{N}(0, \mathbf{I})$ is Gaussian noise, and $t$ is the diffusion timestep. $\epsilon_\theta$ is the learned denoising network. At inference time, the model predicts the entire action trajectory in a denoising pass, and the resulting sequence is executed step-by-step in the environment. The frame–switching mechanism used during inference is described in Appendix A.4.

\subsection{Spatial Reasoning Feedback Policy}
\label{3.3}
To ensure spatial consistency throughout execution, we introduce a Spatial Reasoning Feedback Policy that enables closed-loop correction by dynamically refining the Spatial Plan Table based on perceptual feedback. This module integrates a dual-stage feedback mechanism, operating at both the video validation and action execution levels.

During visual planning, we incorporate a vision-language model (VLM)-based consistency checker inspired by VideoAgent~\cite{videoagent}. A pretrained GPT-4o model is prompted to assess the spatial plausibility of the generated video—verifying object-goal alignment, manipulator trajectories, and overall spatial coherence. Only videos that pass this check are used for downstream control; otherwise, regeneration is triggered under the same spatial condition.

During action execution, we employ rule-based policy diagnostics to monitor temporal progress and detect failures such as positional drift or behavioral stagnation. When triggered, the system computes an updated spatial offset between the end-effector and the target object based on the latest observation. This offset is re-encoded into a refined spatial subplan via GPT-4o, forming an updated condition to replan both the video and action policy.

Through this real-time refinement of spatial intent and control guidance, our system maintains alignment between visual imagination and robotic execution, enabling robust long-horizon manipulation.
\section{Experiments}
\begin{table*}[t]
\centering
\caption{Comparison of task success rates (\%) across 11 MetaWorld tasks. Each method is evaluated over 800 episodes per task. \textbf{R} denotes variants equipped with a replanning mechanism. \textbf{OR} indicates methods that combine online reinforcement learning with replanning. Bold indicates best.}
\label{tab:ours-vs-baselines}
\vspace{-2mm}

\renewcommand{\arraystretch}{1} 
\setlength{\tabcolsep}{3pt}

\resizebox{0.65\textwidth}{!}{
\begin{tabular}{>{\centering\arraybackslash}m{2.5cm}cccccc}
\toprule
\textbf{Method} & door-open & door-close & basketball & shelf-place & btn-press & btn-press-top \\
\midrule
AVDC~\cite{avdc} & 30.7\% & 28.0\% & 21.3\% & 8.0\% & 34.7\% & 17.3\% \\
AVDC-R~\cite{avdc} & 72.0\% & 89.3\% & 37.3\% & 18.7\% & 60.0\% & 24.0\% \\
VA~\cite{videoagent} & 40.0\% & 29.3\% & 13.3\% & 9.3\% & 38.7\% & 18.7\% \\
VA-OR~\cite{videoagent} & 82.7\% & 97.3\% & 40.0\% & 26.7\% & 73.3\% & 44.0\% \\
\midrule
SP & \textbf{100.0\%} & \textbf{100.0\%} & \textbf{77.3\%} & 78.7\% & 96\% & \textbf{98.6\%} \\
SP-R & 98.6\% & \textbf{100.0\%} & 73.3\% & \textbf{92.0\%} & 96.0\% & \textbf{98.6\%} \\
\end{tabular}
\vspace{-2mm}
}

\resizebox{0.65\textwidth}{!}{
\begin{tabular}{>{\centering\arraybackslash}m{2.5cm}cccccc}
\toprule
\textbf{Method} & faucet-close & faucet-open & handle-press & hammer & assembly & \cellcolor{gray!20}\textbf{Average} \\
\midrule
AVDC~\cite{avdc} & 12.0\% & 17.3\% & 41.3\% & 0.0\% & 5.3\% & \cellcolor{gray!20}19.6\% \\
AVDC-R~\cite{avdc} & 53.3\% & 24.0\% & 81.3\% & 8.0\% & 6.7\% & \cellcolor{gray!20}43.1\% \\
VA~\cite{videoagent} & 46.7\% & 12.0\% & 36.0\% & 0.0\% & 1.3\% & \cellcolor{gray!20}22.3\% \\
VA-OR~\cite{videoagent} & 74.7\% & 46.7\% & 86.7\% & 8.0\% & 10.7\% & \cellcolor{gray!20}53.7\% \\
\midrule
SP & \textbf{94.6\%} & 74.7\% & \textbf{100.0\%} & 61.3\% & 21.3\% & \cellcolor{gray!20}82\% \\
SP-R & 92.0\% & \textbf{96.0\%} & \textbf{100.0\%} & \textbf{73.3\%} & \textbf{32.0\%} & \cellcolor{gray!20}\textbf{86.7\%} \\
\bottomrule
\end{tabular}
}
\vspace{-3mm}

\end{table*}

\begin{table}[t]
\centering
\caption{Comparison of task success rates (\%) across 12 tasks in four rooms on iTHOR.}
\label{tab:ithor}
\vspace{-2mm}
\resizebox{0.9\linewidth}{!}{
\renewcommand{\arraystretch}{1}
\setlength{\tabcolsep}{4.5pt} 

\begin{tabular}{c c c c}
\toprule
\textbf{Room} & AVDC~\cite{avdc} & VideoAgent~\cite{videoagent} & SP (Ours) \\
\midrule
Kitchen      & 26.7\% & 28.3\% & \textbf{88.3\%} \\
Living Room  & 23.3\% & 26.7\% & \textbf{50.0\%} \\
Bedroom      & 38.3\% & 41.7\% & \textbf{63.3\%} \\
Bathroom     & 36.7\% & \textbf{40.0\%} & 36.7\% \\
\midrule
\textbf{Average} & 31.3\% & 34.2\% & \textbf{59.6\%} \\
\bottomrule
\end{tabular}
}
\vspace{-4mm}
\end{table}

In this section, we comprehensively evaluate our method across both manipulation and navigation tasks. Sec.~\ref{4.1} introduces the environments and datasets, and Sec.~\ref{4.2} reports the main results on Meta-World and iTHOR. Sec.~\ref{4.3} presents the real-world evaluation results. We further conduct five ablation studies in Sec.~\ref{4.4}, examine failure modes in Sec.~\ref{4.5}, analyze computational efficiency in Sec.~\ref{4.2}, assess robustness to occlusions in Sec.~\ref{4.6}, and study key hyperparameters in Sec.~\ref{4.7}.

\subsection{Datasets and Experimental Setups}
\label{4.1}

\noindent \textbf{Datasets and Environments.} We follow the evaluation setup of \cite{avdc} and conduct experiments in both Meta-World~\cite{metaworld} and iTHOR~\cite{kolve2017ai2thor}. Meta-World is a simulated robotic manipulation benchmark consisting of 11 single-arm control tasks (e.g., pushing, pulling, grasping, inserting); we run 25 random seeds across 11 tasks and 3 camera views, yielding over 800 rollouts. iTHOR is an interactive 3D household environment featuring four room types and 12 navigation tasks, where each task is evaluated with 20 random seeds to assess robustness and cross-scene generalization. Complete training details are provided in Appendix A.2, generated results in Appendix A.3, and dataset details in Appendix A.4.

\noindent\textbf{Baselines and Our Methods}
\begin{itemize}
    \item \textbf{AVDC}: Generates videos via a pretrained diffusion model, extracts optical-flow motion, and inversely infers actions.
    \item \textbf{AVDC-Replan}: Adds replanning by regenerating videos when failures or stagnation occur.
    \item \textbf{VideoAgent}: Refines video plans with a self-supervised consistency rule and selects candidates via GPT-4 scoring.
    \item \textbf{VideoAgent-online-Replan}: Online-updates video generation and optimization using successful trajectories and triggers replanning upon failures.
\end{itemize}

\begin{itemize}
    \item \textbf{SP (Ours)}: Extracts structured spatial plans with a VLM and injects them as global conditions into the diffusion model; actions are derived via a goal-conditioned policy.
    \item \textbf{SP-Replan (Ours)}: Adds iterative regeneration and replanning when execution stagnation is detected.
\end{itemize}

\subsection{Main Results}
\label{4.2}

\paragraph{Meta-World}
As shown in Table~\ref{tab:ours-vs-baselines}, our method achieves an overall success rate of 86.7\%, outperforming all prior baselines. Approaches such as AVDC and VA rely on task text or pretrained flow to infer motion, making them brittle under execution drift and forcing explicit replanning when deviations occur. In contrast, our spatial-plan-driven framework produces structured, goal-aligned video predictions and actions that inherently adapt to spatial variations, resulting in more stable and consistent execution.
On the subset of challenging tasks (\textit{Assembly, Hammer, Faucet Open, Button Press Top, Shelf Place, Basketball}), our method reaches 77.5\% success, whereas the best baseline achieves only 29.2\%, corresponding to a 165.4\% relative improvement. Easier tasks are solved nearly perfectly, often approaching 100\% success.

\begin{figure}[t]
    \centering
    \vspace{-5mm}
    \includegraphics[width=1\linewidth]{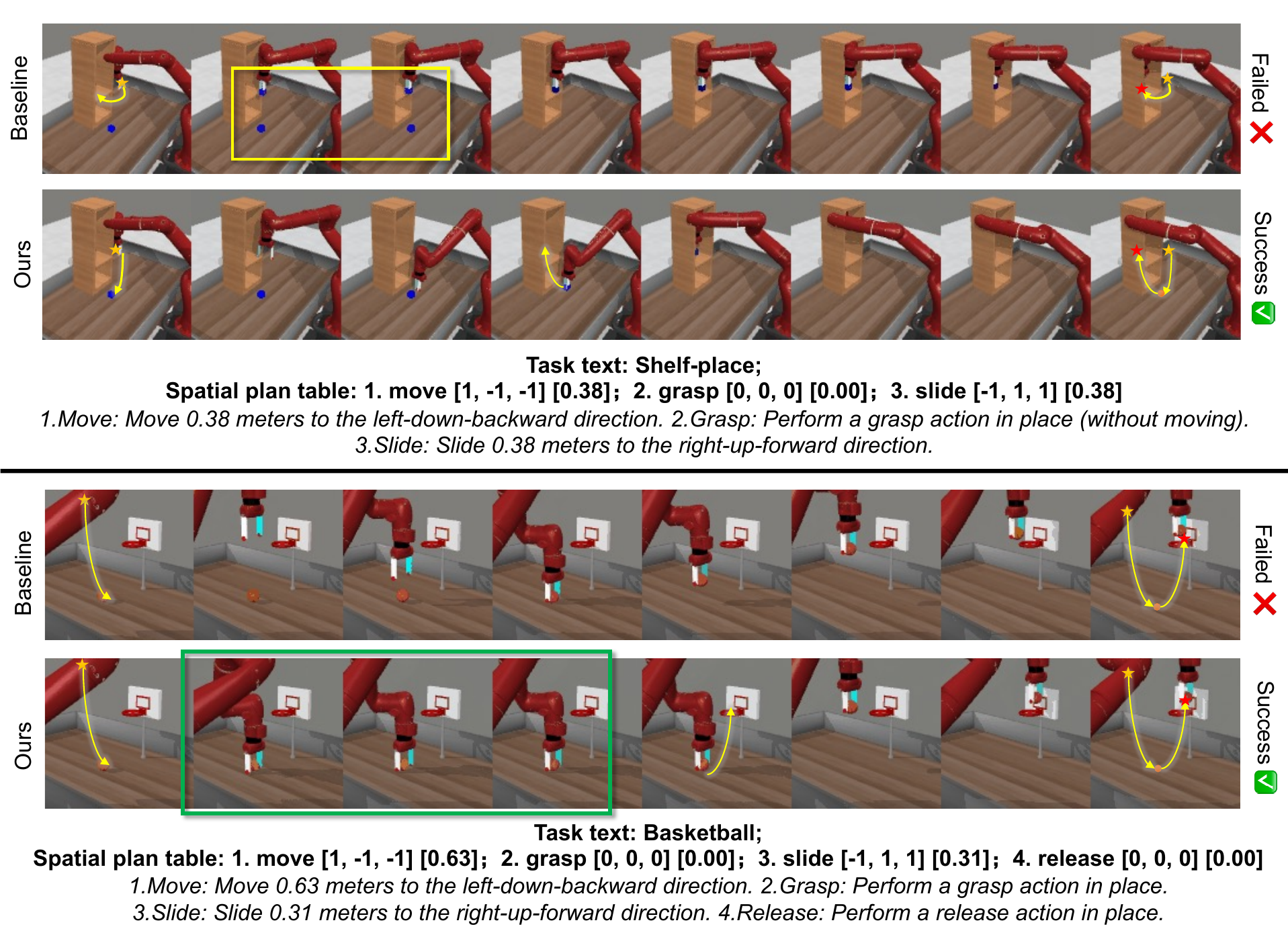}
    \vspace{-7mm}
    \caption{Comparisons on \textit{Shelf Place} and \textit{Basketball} in Meta-World. The yellow line shows the motion trajectory;
    \textcolor{yellow}{\ding{72}} marks the start,
    \textcolor{red}{\ding{72}} the final position,
    and \textcolor{orange}{\ding{108}} intermediate steps.}
    \label{fig:generate}
    \vspace{-4mm}
\end{figure}

\begin{figure}[t]
    \centering
    \includegraphics[width=1\linewidth]{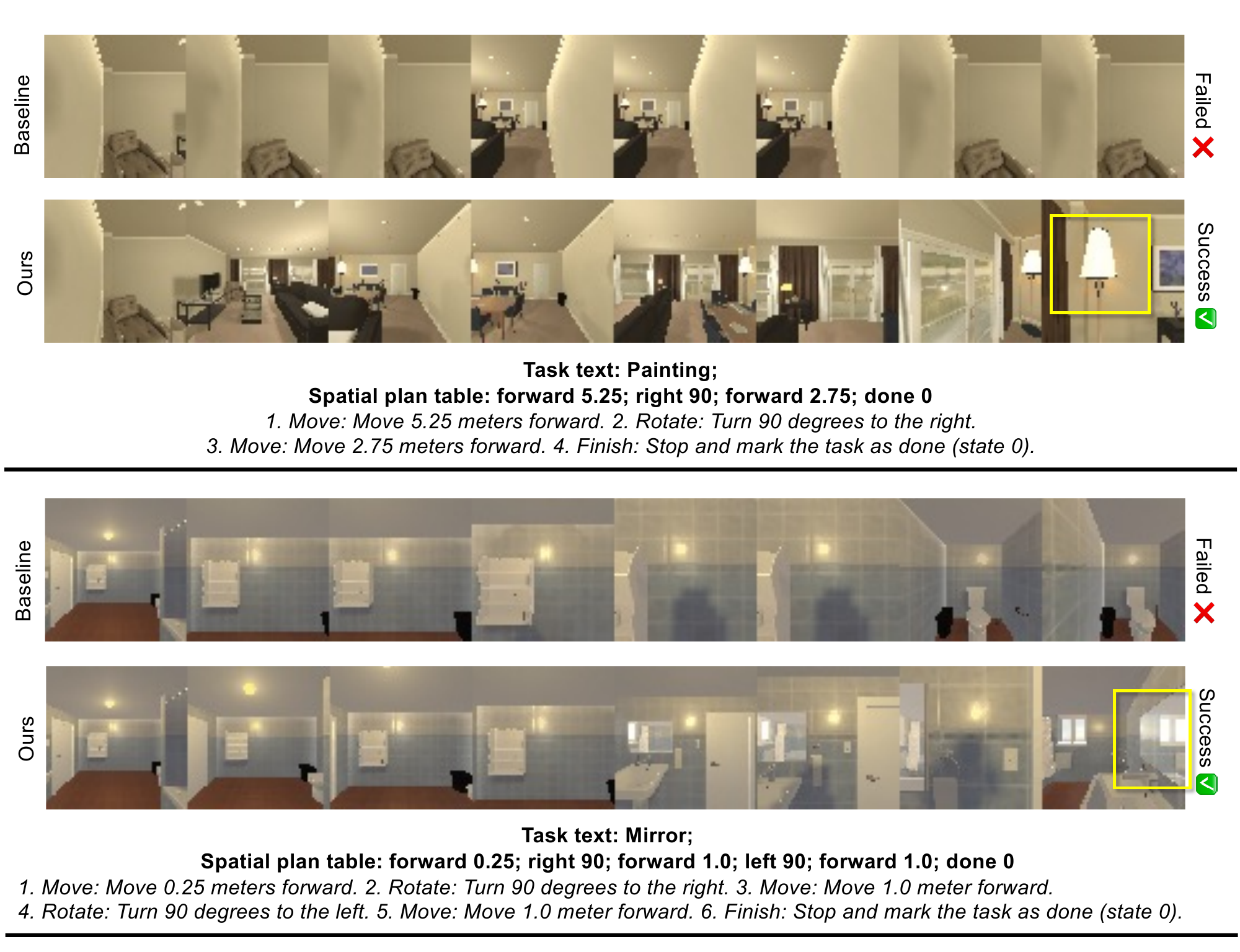}
    \vspace{-7mm}
    \caption{Comparisons on \textit{Painting} and \textit{Mirror}  in iThor. The \textcolor{yellow} {bounding box} marks the target object.}
    \label{fig:generate2}
    \vspace{-4mm}
\end{figure}

Fig.~\ref{fig:generate} illustrates these advantages. In \textit{Shelf Place}, the baseline produces an inflexible trajectory that cannot correct for scene variations, whereas our model generates adaptive, spatially coherent paths. In \textit{Basketball}, our method allocates more frames to the critical grasping phase (green box), yielding smoother and more reliable manipulation, while baseline methods frequently miss key interaction moments.

\begin{figure}[h]
    \centering
    \includegraphics[width=1\linewidth]{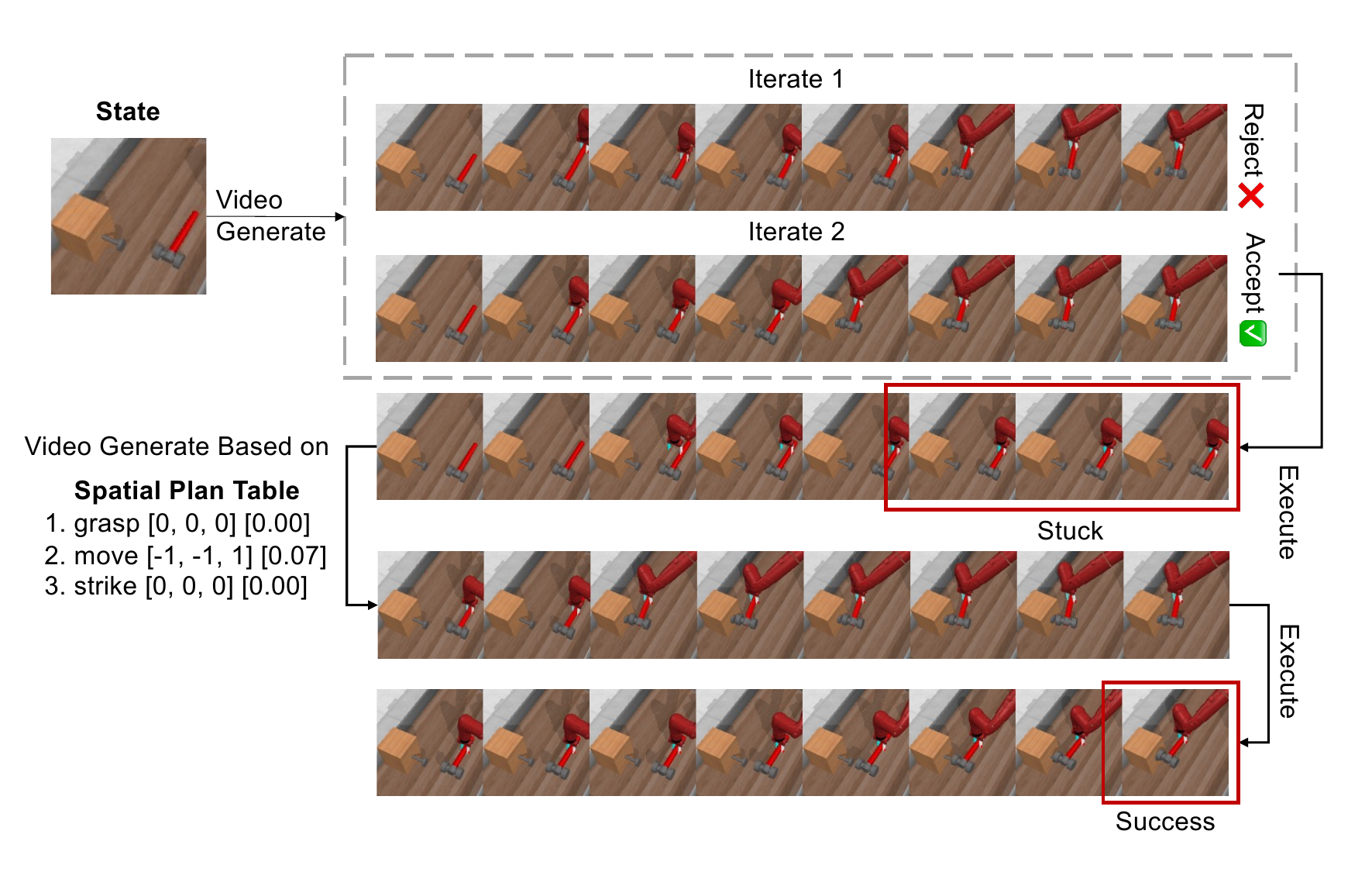}
    \vspace{-7mm}
    \caption{Example of iterative spatial replanning during manipulation in Meta-World.}
    \label{fig:replan}
    \vspace{-4mm}
\end{figure}

Our approach is also efficient. On Meta-World, VLM inference takes 2.43\,s, video generation 9.68\,s, and action execution 0.75\,s, averaging 15\,s per task. The end-to-end runtime is 27.11\,s, compared to 50\,s for VideoAgent, achieving a 1.84$\times$ speedup while delivering higher success rates. Human evaluation further indicates 82\% accuracy in spatial plan table generation and 67\% accuracy in replanning decisions, confirming the interpretability and reliability of our spatial reasoning process.

\paragraph{iTHOR}
Table~\ref{tab:ithor} shows that our method reaches an average success rate of 59.6\% on iTHOR, clearly above the 31.3\% of AVDC and the 34.2\% of VideoAgent. The largest improvements appear in Kitchen with 88.3\% and in Bedroom with 63.3\%, where cluttered and variable layouts require precise reasoning about object positions and feasible motion. Living Room also benefits from stronger spatial grounding, reaching 50.0\%. Bathroom performance is similar to the baselines due to its highly constrained geometry, which limits the impact of planning. Overall, the results indicate that explicit spatial structure, rather than relying solely on visual priors, is crucial for generalizing across diverse room configurations.

The qualitative rollouts in Fig.~\ref{fig:generate2} further illustrate this advantage. Baseline trajectories drift because diffusion priors alone cannot preserve stable spatial alignment, leading to early pose errors in long-horizon tasks such as Painting and Mirror. This issue is amplified in iTHOR, where frequent camera rotations require the model to maintain scene consistency across changing viewpoints. Our method preserves the global layout more faithfully: generated frames remain coherent after rotation, and the target is localized consistently throughout the sequence. By encoding spatial plans that constrain depth reasoning, object–agent relations, and admissible motion, the model produces trajectories that better reflect the actual room topology even under viewpoint shifts.

\subsection{Real-World Experiments}
\label{4.3}

\begin{figure}[t]
  \centering
  \vspace{-5mm}
  \includegraphics[width=\linewidth]{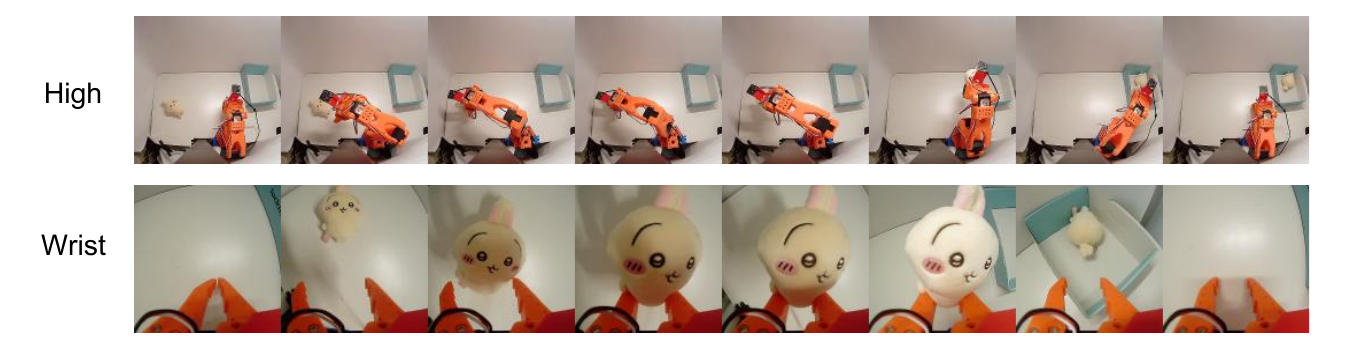}
  \vspace{-7mm}
  \caption{Real-world execution of the \textit{pick doll} task.}
  \label{fig:real_pick_doll}
  \vspace{-5mm}
\end{figure}

We evaluate our system in a real-world setting using an SO-100 robotic arm~\cite{so100_wiki} on a \textit{pick-doll} task, where the robot must grasp a yellow rabbit toy and place it into a box. Since real-world data lack depth, we estimate depth with DepthAnything~\cite{depthanything} and feed the RGB frame, predicted depth, and object spatial information into the spatial plan table. For action prediction, we use the top-down current and goal frames together with the wrist-camera frame to improve spatial grounding. As shown in Figure~\ref{fig:real_pick_doll}, we evaluate 20 trials and achieve a 30\% success rate, \textit{indicating that our model can still complete the task even under incomplete visual observations and depth estimation noise.}

\subsection{Ablation Study}
\label{4.4}

\noindent \textbf{Ablation on the Video Generation Module.}
We evaluate three strategies in the video generation module in Meta-World. (1) The \textit{Baseline} follows AVDC and conditions the diffusion model only on the task prompt and initial frame. (2) Our \textit{Distance} variant adds the spatial offset between the end effector and the target as raw spatial guidance. (3) Our \textit{Subplan} variant transforms this offset into structured action cues using a vision language model and injects them as high level spatial guidance
As shown in Table~\ref{tab:abla1}, both spatial variants outperform the AVDC baseline, demonstrating the \textit{utility of explicitly modeling spatial structures for enhanced visuomotor performance}. Moreover, the subplan-based approach achieves further gains by abstracting raw spatial cues into structured guidance, highlighting the \textit{importance of structured spatial abstraction in facilitating precise and actionable visual plan generation}.

\noindent \textbf{Ablation on the Action Prediction Module.}
We compare two action prediction strategies in Meta-World. (1) AVDC converts optical flow between the current and goal frames into a target coordinate and executes it with a classical controller and heuristic gripper rules. (2) Our method instead uses a flow-based diffusion policy that directly predicts actions from the current frame, goal frame, flow, and robot spatial coordinates, allowing spatial and temporal cues to be modeled jointly.
As shown in Table~\ref{tab:abla2}, our flow-based diffusion policy consistently outperforms AVDC across all difficulty levels. The average success rate rises from 19.6\% for AVDC to 63.4\% for our method, with especially large gains on Medium, Hard, and Very Hard tasks. These results highlight that \textit{learning based spatial reasoning from flow yields more reliable and adaptable action generation than mapping flow fields to target coordinates through hand-designed controllers.}

\noindent \textbf{Ablation on Spatial Condition Injection Strategy in Video Generation.}
We compare two strategies for injecting spatial plans into the diffusion model. (1) The \textit{Global Cond} variant concatenates the spatial plan table with the task embedding and applies it uniformly to all U\!-\!Net layers. (2)The \textit{Local Cond} variant follows the GLIGEN formulation and injects the spatial plan only into intermediate layers through cross attention, enabling more localized spatial modulation.
As shown in Table~\ref{tab:abla3}, Local Conditioning yields higher success rates, reaching 84.3\% on average and improving performance on harder tasks such as Hard and Very Hard. This suggests that \textit{localized spatial injection improves spatial precision in video synthesis and leads to more reliable downstream control.}

\begin{table}[t]
\centering
\caption{Comparison of task success rates (\%) using different video generation strategies across four levels of task difficulty in Meta-World. \textbf{AVDC} uses text and initial frame only; \textbf{Distance} adds raw spatial offsets; \textbf{Subplan} injects structured spatial subplans.}
\label{tab:abla1}
\vspace{-2mm}

\renewcommand{\arraystretch}{1}
\setlength{\tabcolsep}{1.35pt} 
\resizebox{0.85\linewidth}{!}{
\begin{tabular}{c c c c c c}
\toprule
\textbf{} & Easy & Medium & Hard & Very Hard & \textbf{Average} \\
\midrule
AVDC     & 75.0\%      & 48.7\%   & \textbf{26.7\%} & 48.0\%      & 63.4\% \\
Distance & 93.5\%      & 56.7\%   & 25.3\%          & \textbf{81.3\%} & 79.4\% \\
Subplan  & \textbf{94.8\%} & \textbf{69.3\%} & 21.3\%           & 78.7\%        & \textbf{82.0\%} \\
\bottomrule
\end{tabular}
\vspace{-5mm}
}
\end{table}

\begin{table}[t]
\centering
\caption{Comparison of task success rates (\%) across four difficulty levels in Meta-World. \textbf{AVDC} refers to AVDC’s action predictor that directly maps optical flow to actions. \textbf{Ours} uses a flow-based diffusion policy.}
\label{tab:abla2}
\vspace{-2mm}

\renewcommand{\arraystretch}{1}
\setlength{\tabcolsep}{1.35pt} 
\resizebox{0.85\linewidth}{!}{
\begin{tabular}{c c c c c c}
\toprule
\textbf{} & Easy & Medium & Hard & Very Hard & \textbf{Average} \\
\midrule
AVDC & 25.9\% & 10.7\% & 5.3\% & 8.0\% & 19.6\% \\
Ours & \textbf{75.0\%} & \textbf{48.7\%} & \textbf{26.7\%} & \textbf{48.0\%} & \textbf{63.4\%} \\
\bottomrule
\end{tabular}
}
\end{table}

\begin{table}[t]
\centering
\caption{Comparison of task success rates (\%) using different spatial condition injection strategies for video generation across four levels of task difficulty in Meta-World. \textbf{Global Cond} injects the spatial plan table across all U-Net blocks; \textbf{Local Cond} follows GLIGEN-style partial injection via cross-attention at intermediate layers.}
\label{tab:abla3}
\vspace{-2mm}

\renewcommand{\arraystretch}{1}
\setlength{\tabcolsep}{1.35pt} 
\resizebox{0.85\linewidth}{!}{
\begin{tabular}{c c c c c c}
\toprule
\textbf{} & Easy & Medium & Hard & Very Hard & \textbf{Average} \\
\midrule
Global Cond & 94.8\% & \textbf{63.3\%} & 21.3\% & 78.7\% & 82.0\% \\
Local Cond  & \textbf{99.0\%} & 62.0\% & \textbf{29.3\%} & \textbf{80.0\%} & \textbf{84.3\%} \\
\bottomrule
\end{tabular}
}
\end{table}

\noindent \textbf{Ablation on the Replanning Mechanism.}
We compare three replanning settings: (1) \textit{no replanning}, (2)\textit{replanning only during video generation}, and (3) \textit{replanning during both video generation and real execution}. As shown in Table~\ref{tab:abla4}, performance improves as more stages incorporate replanning.
Replanning in both imagination and execution achieves the highest average success rate of 86.7\%, with clear gains on harder tasks. While generation only replanning offers moderate improvement, combining it with execution time replanning enables the system to correct failures, adapt to environmental changes, and stabilize long horizon behavior. Without replanning, the policy cannot recover from unexpected variations. These results indicate that \textit{using replanning in both stages is essential for robust long horizon control.}

\begin{table}[t]
\centering
\caption{Comparison of task success rates (\%) using different replanning mechanisms across four levels of task difficulty in Meta-World. \textbf{NR} denotes no replanning; \textbf{GR} denotes generation replanning; \textbf{ER} denotes execution replanning. Bold indicates best.}
\label{tab:abla4}

\renewcommand{\arraystretch}{1}
\setlength{\tabcolsep}{1.35pt} 
\resizebox{0.85\linewidth}{!}{
\begin{tabular}{c c c c c c}
\toprule
\textbf{} & Easy & Medium & Hard & Very Hard & \textbf{Average} \\
\midrule
NR      & 94.8\% & \textbf{63.3\%} & 21.3\% & 78.7\% & 82.0\% \\
GR      & 96.0\% & 70.7\% & 24.0\% & 80.0\% & 83.4\% \\
GR\&ER  & \textbf{97.3\%} & \textbf{73.3\%} & \textbf{32.0\%} & \textbf{92.0\%} & \textbf{86.7\%} \\
\bottomrule
\end{tabular}
}
\vspace{-5mm}
\end{table}

\noindent \textbf{Ablation on Finetuning with Expert-Assisted Recovery Data.}
We evaluate two training configurations. The \textit{Original} variant uses only fully successful demonstrations, while the \textit{Expert-Assisted} variant is finetuned on additional trajectories where failures are corrected by an expert policy.
As shown in Table~\ref{tab:expert_recovery}, expert assisted finetuning improves performance across all difficulty levels, raising the average success rate from 82.0\% to 85.6\% and yielding substantial gains on Medium and Hard tasks. These results indicate that exposure to partial-to-successful executions enhances robustness by covering recovery scenarios absent in the original dataset. \textit{Overall, expert assisted augmentation provides a simple and effective boost to long-horizon reliability.}

\begin{table}[t]
\centering
\caption{Comparison of task success rates (\%) with and without expert assisted data augmentation across four difficulty levels in Meta-World. \textbf{Original} denotes the model trained only on the standard dataset. \textbf{Expert-Assisted} denotes the model additionally trained on expert-corrected rollouts.}
\label{tab:expert_recovery}
\vspace{-2mm}
\renewcommand{\arraystretch}{1}
\setlength{\tabcolsep}{1.35pt}
\resizebox{0.85\linewidth}{!}{
\begin{tabular}{c c c c c c}
\toprule
\textbf{} & Easy & Medium & Hard & Very Hard & \textbf{Average} \\
\midrule
Original            & 94.8\% & 63.3\% & 21.3\% & 78.7\% & 82.0\% \\
Expert-Assisted     & \textbf{96.5\%} & \textbf{74.9\%} & \textbf{38.7\%} & \textbf{84.0\%} & \textbf{85.6\%} \\
\bottomrule
\end{tabular}
}
\end{table}
\subsection{Error Type Analysis}
\label{4.5}

Human evaluation on MetaWorld reveals that 65\% of the generated videos are correct, with errors mainly attributed to fluency issues (9.1\%), robot inconsistency (11.0\%), and object inconsistency (14.5\%). For action execution, 85.5\% of the predictions are correct, with the remaining errors caused by physics violations (9.1\%) and spatial mistakes (5.5\%). The low spatial error rate indicates that \textit{the spatial planning module and spatial-conditioned policy effectively capture task geometry}, and most failures arise from challenging contact dynamics rather than misinterpreted spatial relations.

\subsection{Robustness to Occlusions via Masking}
\label{4.6}

\begin{figure}[t]
  \vspace{-2mm} 
  \centering
  \includegraphics[width=\linewidth]{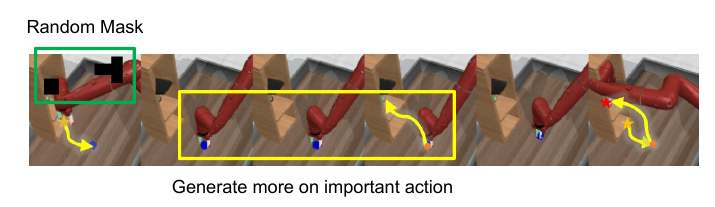}
  \vspace{-7mm} 
  \caption{Video generation results with partial occlusion by random masking. 
  Despite masked regions (left), the model generates coherent future frames (right) and maintains correct task execution.}
  \label{fig:mask_generation}
  \vspace{-5mm} 
\end{figure}

To evaluate the robustness of our video generation module under partial visual occlusions, we introduce a masking mechanism before feeding the current frame into the diffusion model. Specifically, we randomly mask a portion of the image input, simulating occlusions such as those caused by environmental factors or sensor noise.

As illustrated in Figure~\ref{fig:mask_generation}, although significant areas of the current frame are masked, the generated frames remain visually consistent and the downstream agent successfully completes the manipulation tasks. This indicates that the model has learned a robust representation of both visual and temporal context, and can infer plausible continuations even when critical input regions are missing.
\subsection{Hyperparameter Analysis}
\label{4.7}

\begin{figure}[t]
    \vspace{-5mm}
    \centering
    \includegraphics[width=0.9\linewidth]{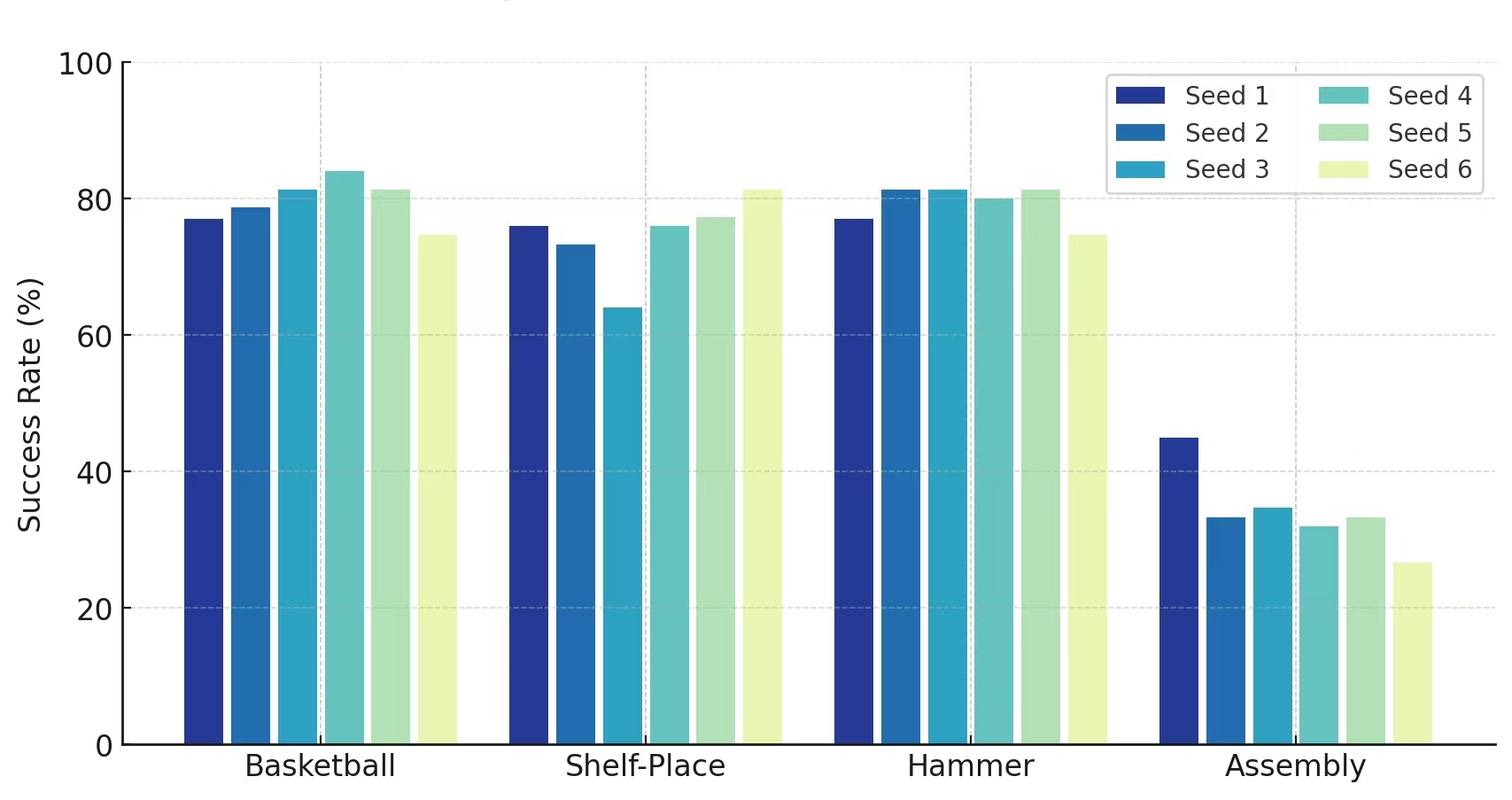}
    \vspace{-5mm}
    \caption{Task success rates on four challenging MetaWorld tasks under different diffusion noise seeds using our SP method, demonstrating consistent performance across seeds.}
    \label{fig:seed}
    \vspace{-5mm}
\end{figure}

\noindent \textbf{Effect of Diffusion Noise Seed in Video Generation.}
To assess the robustness of our video generation module under different stochastic conditions, we conduct experiments using our proposed method (SP) on four challenging MetaWorld tasks: \textit{basketball}, \textit{shelf-place}, \textit{hammer}, and \textit{assembly}. For each task, we sample six different diffusion noise seeds and report the task success rates. As shown in Figure~\ref{fig:seed}, our approach consistently achieves high performance across seeds, with an average success rate around 77\% and low variance. This indicates that the structured spatial subplans effectively anchor the generation process, ensuring the production of reliable and executable visual trajectories even under varying random conditions.

\noindent \textbf{Effect of Hyperparameter in Action Prediction.}
We analyze key design choices in the action prediction module. First, we study the effect of goal frame selection range. Setting the goal frame as a random future frame within the next $k = 20$ timesteps yields over 10\% higher average success rates compared to restricting it to a narrower window between 17–20 timesteps ahead, indicating that allowing a broader temporal horizon improves generalization and task adaptability. Second, regressing the noise $\epsilon$ achieves better performance than directly predicting the denoised action $\hat{a}_0$, which contrasts with some prior findings~\cite{diffusionpolicy}. Lastly, while direct MLP-based action regression converges faster, the diffusion-based denoising model demonstrates superior performance and robustness across tasks.
\section{Conclusion}
We present \textbf{SP}, a spatial-aware visuomotor framework that integrates explicit spatial modeling and reasoning into robot manipulation. Central to our approach is a spatial plan table that encodes atomic actions, directions, and distances, enabling structured spatial guidance throughout the pipeline. Our policy generates spatially coherent visual trajectories by conditioning a video diffusion model on a spatial plan table, which are then used to guide action prediction based on temporally aligned spatial coordinates.
A dual-stage replanning mechanism enhances spatial reasoning by refining the spatial plan table and regenerating video and actions upon inconsistency, ensuring closed-loop correction. Experiments on MetaWorld show that SP improves success rates by over 33\% compared to prior methods, validating the effectiveness of spatial modeling and reasoning.
\clearpage
\setcounter{page}{1}
\maketitlesupplementary
\setcounter{section}{0}
\renewcommand{\thesection}{A.\arabic{section}}

\section{Inference Pipeline}
\label{sec:a1}
As shown in alg.~\ref{alg:inf}, Our inference process consists of three stages: spatial-conditioned embodied video generation, spatial-based action prediction, and spatial reasoning feedback policy. Given an initial observation and the current spatial coordinates of the robot and object, we compute a relative spatial offset $\Delta \mathbf{p}$, which is used to generate a spatial plan table via a VLM. The spatial plan table is fused with the task prompt and fed into a diffusion video generator to produce a goal-directed visual trajectory. 

To infer executable actions, the action prediction model takes the predicted video sequence and encodes the current spatial state, represented solely by the robot end effector coordinate $\mathbf{p}_{\text{ee}}$, through a learnable MLP. This enables precise grounding of actions in the spatial context. During execution, a feedback module monitors task progress. If stagnation is detected, a two stage replanning mechanism is triggered: the spatial plan table is refined, and a new video and action sequence are generated to resume progress. This pipeline enables real time, spatially consistent decision making under dynamic conditions.

\begin{algorithm*}[t]
\caption{\textbf{Inference Pipeline}}
\label{alg:inf}
\KwIn{Environment $\mathcal{E}$, Task prompt $\mathcal{T}_0$, Policy $\pi$}
\KwOut{Success flag $\mathcal{S}$}

\ensuremath{\mathbf{I}_0 \leftarrow \mathcal{E}.\texttt{Render}(),\;
(\mathbf{p}_{\text{ee}},\mathbf{p}_{\text{obj}}) \leftarrow \mathcal{E}.\texttt{GetSpatialState}()} \;

\ensuremath{\mathcal{S}_0 \leftarrow \texttt{GPT4o}(\Delta\mathbf{p}=\mathbf{p}_{\text{obj}}-\mathbf{p}_{\text{ee}}),\;
\mathcal{T}\leftarrow[\mathcal{T}_0;\mathcal{S}_0]} \;

\Repeat{\ensuremath{\texttt{VLMCheck}(\mathbf{V}_{1:T})} succeeds}{
  \ensuremath{\mathbf{V}_{1:T}\leftarrow \texttt{DiffusionGenerate}(\mathbf{I}_0,\mathcal{T})}}
  
\ensuremath{\mathbf{z}_{\text{coord}}\leftarrow \texttt{EncodeCoord}(\mathbf{p}_{\text{ee}},\mathbf{p}_{\text{obj}}),\quad
\mathcal{A}_{1:H}\leftarrow \pi(\mathbf{V}_{1:T},\mathbf{z}_{\text{coord}})} \;

\For{\ensuremath{t=1} \KwTo \ensuremath{H}}{
  \ensuremath{\mathcal{E}.\texttt{Step}(\mathcal{A}_t)} \;
  \If{\texttt{IsStuck}(\ensuremath{\mathcal{E}})}{
    \ensuremath{\mathbf{I}_t\leftarrow \mathcal{E}.\texttt{Render}(),\;
    (\mathbf{p}_{\text{ee}},\mathbf{p}_{\text{obj}})\leftarrow \mathcal{E}.\texttt{GetSpatialState}(),\;
    \mathcal{S}_t\leftarrow \texttt{GPT4o}(\Delta\mathbf{p}),\;
    \mathcal{T}_t\leftarrow [\mathcal{T}_0;\mathcal{S}_t]} \;
    \Repeat{\ensuremath{\texttt{VLMCheck}(\mathbf{V}_{1:T})} succeeds}{
      \ensuremath{\mathbf{V}_{1:T}\leftarrow \texttt{DiffusionGenerate}(\mathbf{I}_t,\mathcal{T}_t)}}
    \ensuremath{\mathbf{z}_{\text{coord}}\leftarrow \texttt{EncodeCoord}(\mathbf{p}_{\text{ee}},\mathbf{p}_{\text{obj}}),\quad
    \mathcal{A}_{1:H}\leftarrow \pi(\mathbf{V}_{1:T},\mathbf{z}_{\text{coord}})} \;
  }
}

\ensuremath{\mathcal{S} \leftarrow \mathcal{E}.\texttt{CheckSuccess}()} \;
\Return{\ensuremath{\mathcal{S}}}
\end{algorithm*}

\section{Training Details: Video Generation and Action Prediction}
\label{sec:a2}

\subsection{Video Generation}
\label{app:videogenerate}

\paragraph{Training Settings.}  
All models are trained on $128 \times 128$ RGB frames sampled as 8-frame clips. We use the AdamW optimizer~\cite{adam} with an initial learning rate of $1 \times 10^{-4}$ and a cosine learning rate schedule. Training is conducted for 100{,}000 steps, unless otherwise specified. The standard batch size is 8 for training and 16 for validation, with gradient accumulation set to 1. Mixed precision training is employed using FP16 and automatic mixed precision (AMP). A linear diffusion noise schedule is used with $\beta_1 = 1 \times 10^{-4}$ and $\beta_T = 0.02$ over 1000 steps. Exponential Moving Average (EMA) with a decay factor of 0.999 is updated every 10 training steps. Conditional dropout with a probability of 0.1 is applied to all models except the Baseline. Validation uses 1 or 2 samples per batch depending on the script. Unless otherwise specified, all training datasets are collected and generated from our own single successful executions. All experiments are conducted on a single NVIDIA RTX 4090 GPU (80GB memory), with full-precision (FP32) training completed within 2 days.

\paragraph{Inference.}  
During inference, all models use DDIM sampling with classifier-free guidance. Models conditioned on plans demonstrate improved semantic consistency and object interaction fidelity.

\paragraph{Model Variants and Conditioning.}  
\label{app:differentmodelsetting }
We investigate four variants of our conditional diffusion models, differing mainly in conditioning inputs and architectural enhancements. Key differences and training datasets are summarized in Table~\ref{tab:video_gen_versions}.

\begin{table}[ht]
\centering
\caption{Comparison of video generation model variants highlighting key architectural or training differences.}
\label{tab:video_gen_versions}

\resizebox{0.48\textwidth}{!}{%
\begin{tabular}{p{0.5\linewidth}|p{0.9\linewidth}}
\toprule
\textbf{Version} & \textbf{Description} \\ \midrule

Baseline (Global Conditioning on Task) & 
Standard UNet conditioned on global CLIP text embedding, serving as a task-aware video prior. \\[6pt]
\midrule

Global Conditioning on Task \& Distance &
Adds a 9-dimensional gripper-object distance vector encoder for enhanced spatial grounding. \\[6pt]
\midrule

Global Conditioning on Task \& Subplan &
Incorporates structured multi-step action plans as global conditioning signals. \\[6pt]
\midrule

Global Conditioning on Task \& Subplan (Fine-tuned) &
Initially trained on expert data for 10k steps, then finetuned on additional datasets for 2k steps. \\[6pt]
\midrule

Local Conditioning on Task \& Subplan &
Introduces cross-attention to fuse global and local plan features for fine-grained temporal conditioning. \\
\bottomrule
\end{tabular}
}
\end{table}

\subsection{Action Prediction}
\paragraph{Training Settings.}
Unlike the original Diffusion Policy~\cite{diffusionpolicy}, which conditions on past observations and actions, our model relies solely on the current observation, future observation, and spatial context. Specifically, instead of using a standard 3-channel RGB input, we concatenate the current frame and a sampled goal frame to form a 6-channel input image of resolution $128 \times 128$, along with spatial coordinates. Given horizon $H=4$, the model predicts $H$ future actions from a single observation step ($T=1$), conditioned on the current frame, spatial coordinates, and a goal frame sampled within the next $K=20$ frames. 
We adopt a ResNet-50 visual encoder~\cite{resnet} pretrained on ImageNet~\cite{imagenet}, which remains trainable during training. The spatial coordinates are encoded via a MLP, which is also trained end-to-end. The policy network is implemented as a conditional U-Net with diffusion denoising, using a cosine noise schedule ($\beta_1 = 1 \times 10^{-4}$, $\beta_T = 2 \times 10^{-2}$) over 100 steps. 
The model is optimized using AdamW with learning rate $\eta = 1 \times 10^{-4}$, weight decay $1 \times 10^{-6}$, and batch size $B = 512$ for both training and validation. The learning rate follows a cosine annealing schedule with $500$ warm-up steps. Exponential Moving Average (EMA) is applied with a maximum decay rate of $0.9999$ to stabilize training. All experiments are conducted on a single NVIDIA RTX 4090 GPU (40GB memory), with full-precision (FP32) training completed within 3 days.

\paragraph{Inference.}
During inference, we adopt DDIM sampling with classifier-free guidance to generate action sequences. Conditioning on spatial plans significantly enhances semantic alignment and spatial fidelity, leading to more coherent and goal-directed behaviors.

\section{Tasks in Meta-World and iThor}
\label{sec:a3}

\begin{figure}[t]
    \centering
    \includegraphics[width=0.95\linewidth]{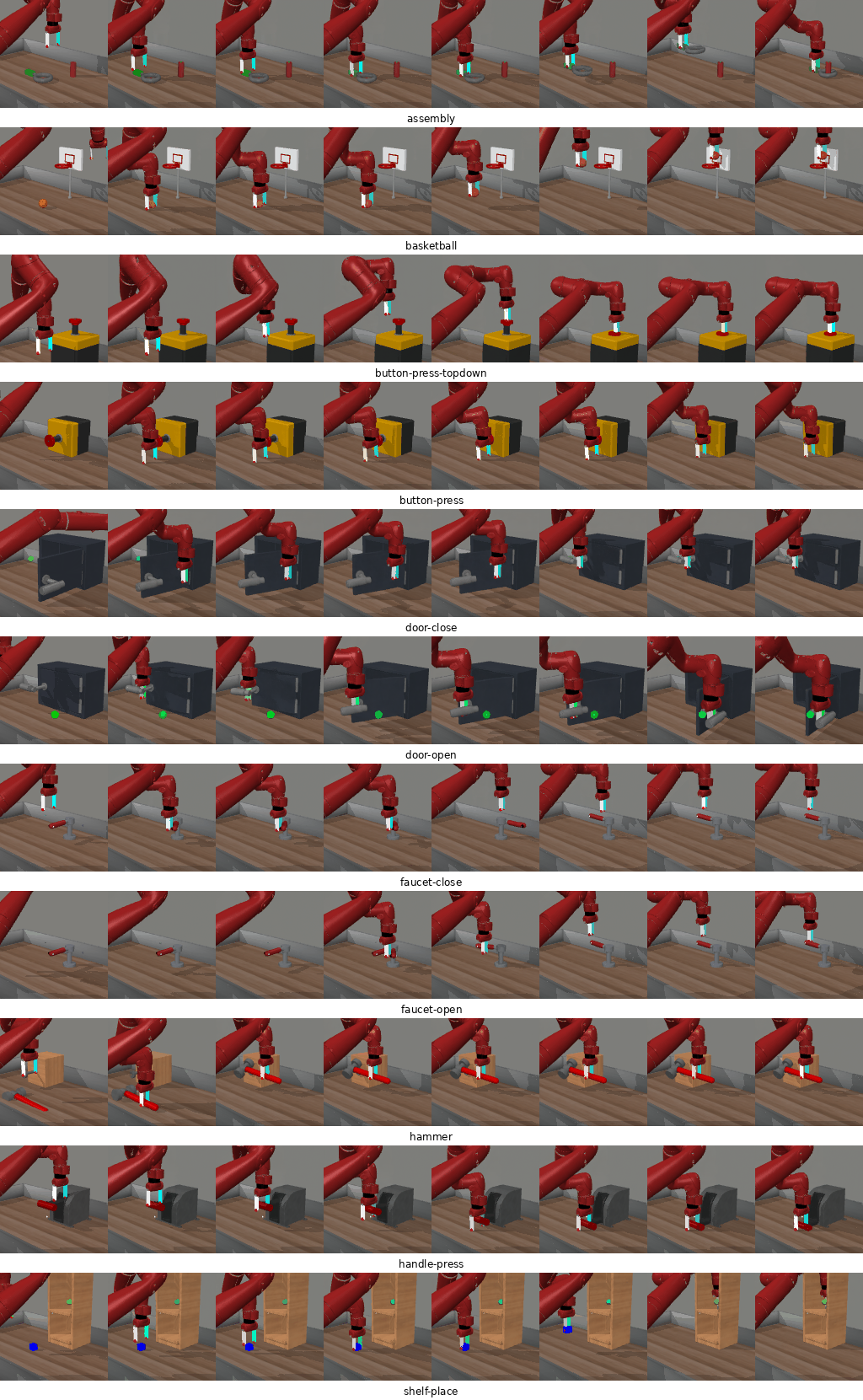}
    \caption{Generated video plans on the 11 Meta-World tasks.}
    \label{fig:mw_generation}
\end{figure}

\noindent \textbf{Meta-World.}
In the Meta-World benchmark~\cite{metaworld}, we evaluated the performance on 11 representative robotic manipulation tasks shown in Figure \ref{fig:mw_generation}, which are designed to assess various manipulation skills and generalization abilities of reinforcement learning algorithms. These tasks are categorized by their difficulty levels as follows:

\begin{itemize}
    \item \textbf{Very Hard:}
    \textit{Shelf Place}: The agent must grasp an object and place it onto the target shelf. This task is particularly challenging due to the precision required in both grasping and placing the object accurately on the shelf.
    \item \textbf{Hard:}
    \textit{Assembly}: The agent grasps the handle position and inserts the ring into the slot to complete assembly. This task requires precise manipulation and coordination of multiple parts.
    \item \textbf{Medium:}
    \textit{Hammer}: The agent grasps the hammer handle and strikes forward to push the nail in with the hammer head. This task involves force control and spatial accuracy.
    \textit{Basketball}: The task involves grasping the basketball, lifting it up, moving forward above the hoop, and placing it onto the hoop to complete the task. It requires coordinated movement and spatial awareness.
    \item \textbf{Easy:}
    \textit{Door Open}: The agent is required to push into the door handle grip area and then pull to open the door.
    \textit{Door Close}: The agent needs to push into the door handle grip area and then push to close the door.
    \textit{Button Press}: The agent moves horizontally to the button and pushes it in.
    \textit{Button Press Topdown}: The agent moves above the button and presses down vertically.
    \textit{Faucet Close}: The agent moves to the left side of the faucet handle and turns it right to close.
    \textit{Faucet Open}: The agent moves to the right side of the faucet handle and turns it left to open.
    \textit{Handle Press}: The agent grasps the handle and presses it downwards.
\end{itemize}

These tasks cover a range of manipulation skills, including pushing, pulling, grasping, placing, pressing, and turning, and they involve different objects and goals, providing a comprehensive evaluation of the agent's ability to learn and generalize across various manipulation tasks.

\begin{figure}[t]
    \centering
    \includegraphics[width=0.95\linewidth]{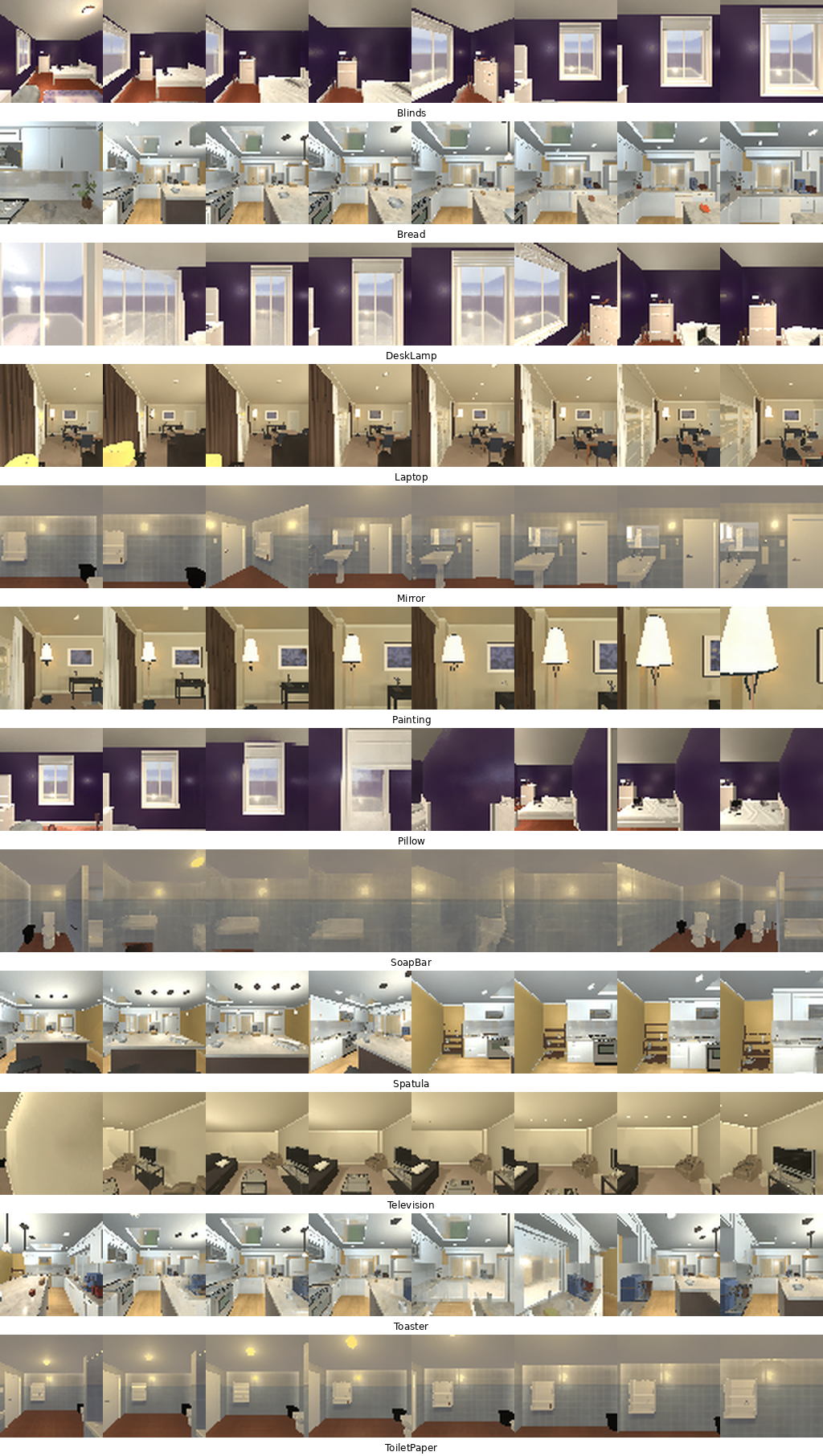}
    \caption{Generated video plans across 12 tasks in four iTHOR room types (Kitchen, Living Room, Bedroom, Bathroom).}
    \label{fig:ithor_generation}
\end{figure}

\noindent \textbf{iThor.}
To assess generalization in visually diverse, embodied environments, we further evaluate on 12 navigation-style manipulation tasks in iTHOR, spanning four rooms: \textit{Kitchen}, \textit{Living Room}, \textit{Bedroom}, and \textit{Bathroom}. Tasks include locating or approaching objects under varying viewpoints and scene layouts, such as \textit{Blinds}, \textit{Bread}, \textit{DeskLamp}, \textit{Laptop}, \textit{Mirror}, \textit{Painting}, \textit{Pillow}, \textit{SoapBar}, \textit{Spatula}, \textit{Television}, \textit{Toaster}, and \textit{ToiletPaper}. The full set of 4 rooms and 12 tasks is shown in Figure~\ref{fig:ithor_generation}.

Together, these two suites provide broad coverage of manipulation and navigation challenges, enabling a comprehensive evaluation of visuomotor generalization across simulated robotic arms and interactive 3D household environments.

\section{Dataset Construction}
\label{sec:a4}

We construct our dataset from expert demonstrations collected in the Meta-World~\cite{metaworld} simulator. For each task, successful execution trajectories are recorded and stored as NPZ and PKL files. Each trajectory contains RGB, depth, and mask images at each timestep, along with task metadata such as environment name, frame index, and camera parameters.

To enhance the usefulness of the dataset for learning fine-grained control, we identify \textit{fine manipulation intervals} using a spatial-aware strategy. Specifically, we compute the 3D positions of both grippers and task-relevant target points by combining mask and rgb-depth information into point clouds. Distances between grippers and targets are monitored across frames to identify close-contact phases indicative of fine manipulation.

Each raw trajectory is temporally segmented, and frames within fine intervals are sampled at 5$\times$ the rate of coarse intervals. The sampling process is governed by the following parameters:

\begin{itemize}
    \item \textbf{Interval:} sampling interval between frames
    \item \textbf{distance\_threshold:} maximum distance to consider a frame as fine-grained
    \item \textbf{consecutive frames:} number of consecutive frames required to enter fine-grained phase
    \item \textbf{recovery needed frames:} number of coarse frames required before re-entering fine phase
    \item \textbf{suppress single spike:} whether to ignore short spikes of distance violations
    \item \textbf{max allowed anomaly:} allowed anomaly frames within fine-grained segment
\end{itemize}

To improve spatial focus, we crop each RGB frame based on task-specific visual layouts. Depending on the task and viewpoint, center or specific regions are retained using manually determined crop windows. The final dataset consists of (1) sampled RGB/mask/depth frames, (2) frame-wise 2D/3D positions of grippers and targets, (3) their pairwise distances, and (4) spatial plan table  in textual form.

\section{VLM Prompt Design}
\label{sec:a5}
\begin{figure*}[h]
    \centering
    \includegraphics[width=1\linewidth]{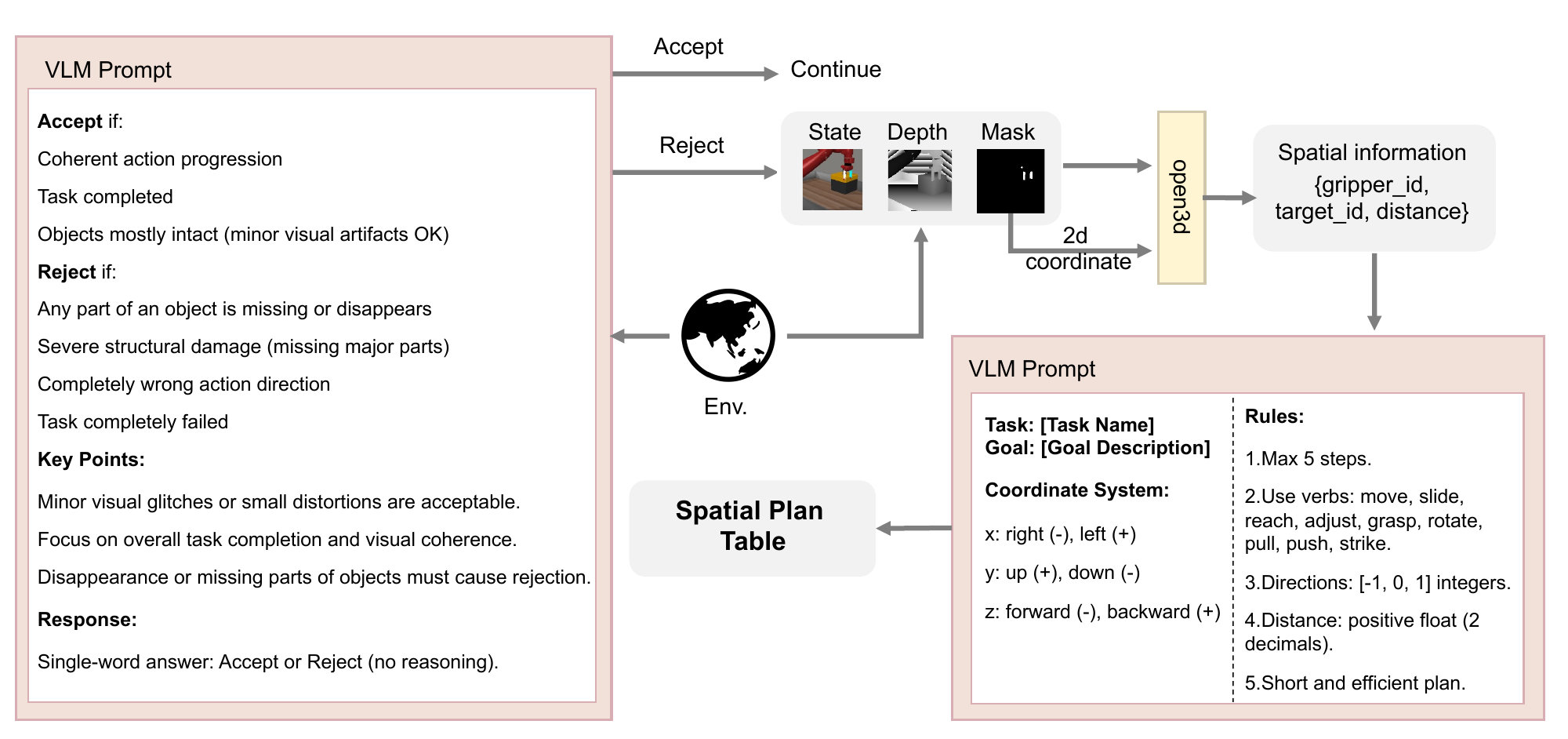}
    \caption{Detailed Design of ours VLM Prompt}
    \label{fig:placeholder1}
\end{figure*}

We explore the use of vision-language models (VLMs) to provide structured feedback in both generation validation and action refinement. While prior works often use open-ended prompts to guide action planning (e.g., “How should the robot complete this task?”), we find that such free-form task descriptions or scene summaries yield inconsistent or inaccurate responses in our setting.

\paragraph{Ineffective Prompt Types (Abandoned):}
\begin{itemize}
    \item \textbf{Task description (ineffective):} Prompts like ``close the door slow'' or ``grasp the basketball carefully'' often fail to produce actionable or reliable feedback due to ambiguity and lack of spatial grounding.
    \item \textbf{Scene summary (ineffective):} Descriptions of the visual scene, such as ``the door is slightly open, and the gripper is nearby'', result in inconsistent VLM interpretation and lack of precise spatial understanding.
\end{itemize}

\paragraph{Effective Prompt Design:}
We instead adopt a \textbf{structured prompt format} tailored to two specific functionalities shown in Fig.~\ref{fig:placeholder1}:

\begin{enumerate}
    \item \textbf{Video validation (generation filtering):}  
    We use a closed-loop prompt to determine whether a generated video should be accepted or rejected. The prompt emphasizes object completeness and task progression, while tolerating minor visual artifacts. The VLM is instructed to respond with a strict ``Accept'' or ``Reject'' based on visual quality. 

    This strict, rule-based instruction significantly improves agreement with human judgments, especially in detecting object disappearance or hallucination.

    \item \textbf{Spatial plan table generation:}  
    As described in Section ~\ref{sec:a4}, we format spatial plan table into structured multi-step plans of the form:

    \begin{lstlisting}
Plan:
1. move [-1, 0, 0] [0.19]
2. move [0, 0, -1] [0.21]
3. push [0, 0, 0] [0.00]
    \end{lstlisting}

Each line encodes an action, a direction vector (x: right-/left+, y: up+/down-, z: forward-/back+), and a positive float distance (in meters). These are generated by aligning frame transitions with gripper/target spatial dynamics and then converted into text for conditioning our VLM-enhanced modules.

\end{enumerate}

\paragraph{Conclusion:}
Our experiments suggest that \textit{free-form prompts are unreliable} for task-level planning or video validation in closed-loop robotic control. Instead, \textbf{spatial plan table with explicit roles, response formats, and spatial semantics} lead to significantly more stable and interpretable VLM outputs, which are essential for real-time feedback in visual policy generation.
\section{Frame Matching Mechanism}
\label{sec:a6}
To ensure precise alignment between visual goals and execution states, we introduce a similarity-based frame matching mechanism. At each timestep, the current observation frame $\mathbf{I}_{\text{cur}}$ is compared to the current target frame $\hat{\mathbf{I}}_g$ sampled from the predicted video sequence $\{\hat{\mathbf{I}}_1, \hat{\mathbf{I}}_2, \ldots, \hat{\mathbf{I}}_K\}$. We compute a composite similarity score via:

\begin{equation}
\begin{split}
\text{Sim}(\mathbf{I}_{\text{cur}}, \hat{\mathbf{I}}_g) = 
&\ w_{\text{geo}} \cdot \text{Sim}_{\text{geo}} + 
   w_{\text{pos}} \cdot \text{Sim}_{\text{pos}} \\
& + w_{\text{ssim}} \cdot \text{SSIM}(\mathbf{I}_{\text{cur}}, \hat{\mathbf{I}}_g) + 
    w_{\text{flow}} \cdot \text{Sim}_{\text{flow}},
\end{split}
\end{equation}

where $\text{Sim}_{\text{geo}}$ is the similarity based on contour and shape features~\cite{lowe2004distinctive}, $\text{Sim}_{\text{pos}}$ encodes block-wise intensity statistics and center-of-mass shifts, $\text{SSIM}$ is the structural similarity index between grayscale frames~\cite{wang2004image}, and $\text{Sim}_{\text{flow}}$ measures the inverse magnitude of optical flow~\cite{farneback2003two}. The weights are set to $w_{\text{geo}}{=}0.35$, $w_{\text{pos}}{=}0.35$, $w_{\text{ssim}}{=}0.2$, and $w_{\text{flow}}{=}0.1$. A frame is considered matched if:
\begin{equation}
\text{Sim}(\mathbf{I}_{\text{cur}}, \hat{\mathbf{I}}_g) \geq \tau,
\end{equation}
with the threshold $\tau{=}0.8$. Upon match, we advance the goal frame to $\hat{\mathbf{I}}_{g+1}$. To prevent deadlocks when similarity fails to exceed the threshold, a hard switch is enforced every $T_{\text{max}}{=}28$ steps.

{
    \small
    \bibliographystyle{ieeenat_fullname}
    \bibliography{main}

@String(CVPR= {IEEE Conf. Comput. Vis. Pattern Recog.})

@String(ICLR = {Int. Conf. Learn. Represent.})

@String(CVPR  = {CVPR})

@String(ICLR  = {ICLR})

@article{rt2,  
 title={RT-2: Vision-Language-Action Models Transfer Web Knowledge to Robotic Control}, 
 author={Brohan, Anthony and Brown, Noah and Chebotar, Yevgen and Chen, Xi and Choromanski, Krzysztof and Ding, Tianli and Driess, Danny and Dubey, Avinava and Finn, Chelsea and Florence, Pete and Fu, Chuyuan and Arenas, MontseGonzalez and Gopalakrishnan, Keerthana and Han, Kehang and Hausman, Karol and Herzog, Alexander and Hsu, Jasmine and Ichter, Brian and Irpan, Alex and Joshi, Nikhil and Julian, Ryan and Kalashnikov, Dmitry and Kuang, Yuheng and Leal, Isabel and Lee, Lisa and Lee, Tsang-WeiEdward and Levine, Sergey and Lu, Yao and Michalewski, Henryk and Mordatch, Igor and Pertsch, Karl and Rao, Kanishka and Reymann, Krista and Ryoo, Michael and Salazar, Grecia and Sanketi, Pannag and Sermanet, Pierre and Singh, Jaspiar and Singh, Anikait and Soricut, Radu and Tran, Huong and Vanhoucke, Vincent and Vuong, Quan and Wahid, Ayzaan and Welker, Stefan and Wohlhart, Paul and Wu, Jialin and Xia, Fei and Xiao, Ted and Xu, Peng and Xu, Sichun and Yu, Tianhe and Zitkovich, Brianna}, 
 language={en-US}, 
year={2024}
 }

@article{gr2,
  title={Gr-2: A generative video-language-action model with web-scale knowledge for robot manipulation},
  author={Cheang, Chi-Lam and Chen, Guangzeng and Jing, Ya and Kong, Tao and Li, Hang and Li, Yifeng and Liu, Yuxiao and Wu, Hongtao and Xu, Jiafeng and Yang, Yichu and others},
  journal={arXiv preprint arXiv:2410.06158},
  year={2024}
}

@article{diffusionpolicy,
  title={Diffusion policy: Visuomotor policy learning via action diffusion},
  author={Chi, Cheng and Xu, Zhenjia and Feng, Siyuan and Cousineau, Eric and Du, Yilun and Burchfiel, Benjamin and Tedrake, Russ and Song, Shuran},
  journal={The International Journal of Robotics Research},
  pages={02783649241273668},
  year={2023},
  publisher={SAGE Publications Sage UK: London, England}
}

@article{dp3d,
  title={3d diffusion policy: Generalizable visuomotor policy learning via simple 3d representations},
  author={Ze, Yanjie and Zhang, Gu and Zhang, Kangning and Hu, Chenyuan and Wang, Muhan and Xu, Huazhe},
  journal={arXiv preprint arXiv:2403.03954},
  year={2024}
}

@article{clover,
  title={Closed-loop visuomotor control with generative expectation for robotic manipulation},
  author={Bu, Qingwen and Zeng, Jia and Chen, Li and Yang, Yanchao and Zhou, Guyue and Yan, Junchi and Luo, Ping and Cui, Heming and Ma, Yi and Li, Hongyang},
  journal={Advances in Neural Information Processing Systems},
  volume={37},
  pages={139002--139029},
  year={2024}
}

@article{avdc,
  title={Learning to act from actionless videos through dense correspondences},
  author={Ko, Po-Chen and Mao, Jiayuan and Du, Yilun and Sun, Shao-Hua and Tenenbaum, Joshua B},
  journal={arXiv preprint arXiv:2310.08576},
  year={2023}
}

@inproceedings{videoagent,
  title={VideoAgent: Self-Improving Video Generation for Embodied Planning},
  author={Soni, Achint and Venkataraman, Sreyas and Chandra, Abhranil and Fischmeister, Sebastian and Liang, Percy and Dai, Bo and Yang, Sherry},
  booktitle={Workshop on Reinforcement Learning Beyond Rewards@ Reinforcement Learning Conference},
  year={2025}
}

@article{susie,
  title={Zero-shot robotic manipulation with pretrained image-editing diffusion models},
  author={Black, Kevin and Nakamoto, Mitsuhiko and Atreya, Pranav and Walke, Homer and Finn, Chelsea and Kumar, Aviral and Levine, Sergey},
  journal={arXiv preprint arXiv:2310.10639},
  year={2023}
}

@article{seer,
  title={Predictive inverse dynamics models are scalable learners for robotic manipulation},
  author={Tian, Yang and Yang, Sizhe and Zeng, Jia and Wang, Ping and Lin, Dahua and Dong, Hao and Pang, Jiangmiao},
  journal={arXiv preprint arXiv:2412.15109},
  year={2024}
}

@article{imagen,
  title={Imagen video: High definition video generation with diffusion models},
  author={Ho, Jonathan and Chan, William and Saharia, Chitwan and Whang, Jay and Gao, Ruiqi and Gritsenko, Alexey and Kingma, Diederik P and Poole, Ben and Norouzi, Mohammad and Fleet, David J and others},
  journal={arXiv preprint arXiv:2210.02303},
  year={2022}
}

@inproceedings{clip,
  title={Learning transferable visual models from natural language supervision},
  author={Radford, Alec and Kim, Jong Wook and Hallacy, Chris and Ramesh, Aditya and Goh, Gabriel and Agarwal, Sandhini and Sastry, Girish and Askell, Amanda and Mishkin, Pamela and Clark, Jack and others},
  booktitle={International conference on machine learning},
  pages={8748--8763},
  year={2021},
  organization={PmLR}
}

@inproceedings{farneback,
  title={Two-frame motion estimation based on polynomial expansion},
  author={Farneb{\"a}ck, Gunnar},
  booktitle={Scandinavian conference on Image analysis},
  pages={363--370},
  year={2003},
  organization={Springer}
}

@article{spatialvlm,
  title={SpatialVLM: Endowing Vision-Language Models with Spatial Reasoning Capabilities}, 
  author={Chen, Boyuan and Xu, Zhuo and Kirmani, Sean and Ichter, Brian and Driess, Danny and Florence, Pete and Sadigh, Dorsa and Guibas, Leonidas and Xia, Fei},
  journal={arXiv preprint arXiv:2401.12168},
  year={2024},
  url={https://arxiv.org/abs/2401.12168}
}

@article{graspcorrect,
  title={GraspCorrect: Robotic Grasp Correction via Vision-Language Model-Guided Feedback},
  author={Lee, Sungjae and Hong, Yeonjoo and Kim, Kwang In},
  journal={arXiv preprint arXiv:2503.15035},
  year={2025},
  url={https://arxiv.org/abs/2503.15035}
}

@article{3dvla,
  title={3D-VLA: A 3D Vision-Language-Action Generative World Model},
  author={Zhen, Haoyu and Qiu, Xiaowen and Chen, Peihao and Yang, Jincheng and Yan, Xin and Du, Yilun and Hong, Yining and Gan, Chuang},
  journal={arXiv preprint arXiv:2403.09631},
  year={2024},
  url={https://arxiv.org/abs/2403.09631}
}

@article{llmplanner,
  title={LLM-Planner: Few-Shot Grounded Planning for Embodied Agents with Large Language Models},
  author={Song, Chan Hee and Wu, Jiaman and Washington, Clayton and Sadler, Brian M. and Chao, Wei-Lun and Su, Yu},
  journal={arXiv preprint arXiv:2212.04088},
  year={2023},
  url={https://arxiv.org/abs/2212.04088}
}

@article{spa,
  title={SPA: 3D Spatial-Awareness Enables Effective Embodied Representation},
  author={Zhu, Haoyi and Yang, Honghui and Wang, Yating and Yang, Jiange and Wang, Limin and He, Tong},
  journal={arXiv preprint arXiv:2410.08208},
  year={2025},
  url={https://arxiv.org/abs/2410.08208}
}

@inproceedings{metaworld,
  title={Meta-world: A benchmark and evaluation for multi-task and meta reinforcement learning},
  author={Yu, Tianhe and Quillen, Deirdre and He, Zhanpeng and Julian, Ryan and Hausman, Karol and Finn, Chelsea and Levine, Sergey},
  booktitle={Conference on robot learning},
  pages={1094--1100},
  year={2020},
  organization={PMLR}
}

@article{spvla,
  title={SP-VLA: A Joint Model Scheduling and Token Pruning Approach for VLA Model Acceleration},
  author={Li, Ye and Meng, Yuan and Sun, Zewen and Ji, Kangye and Tang, Chen and Fan, Jiajun and Ma, Xinzhu and Xia, Shutao and Wang, Zhi and Zhu, Wenwu},
  journal={arXiv preprint arXiv:2506.12723},
  year={2025}
}

@article{grmg,
  title={Gr-mg: Leveraging partially-annotated data via multi-modal goal-conditioned policy},
  author={Li, Peiyan and Wu, Hongtao and Huang, Yan and Cheang, Chilam and Wang, Liang and Kong, Tao},
  journal={IEEE Robotics and Automation Letters},
  year={2025},
  publisher={IEEE}
}

@article{unipi,
  title={Learning universal policies via text-guided video generation},
  author={Du, Yilun and Yang, Sherry and Dai, Bo and Dai, Hanjun and Nachum, Ofir and Tenenbaum, Josh and Schuurmans, Dale and Abbeel, Pieter},
  journal={Advances in neural information processing systems},
  volume={36},
  pages={9156--9172},
  year={2023}
}

@article{wang2004image,
  title={Image quality assessment: from error visibility to structural similarity},
  author={Wang, Zhou and Bovik, Alan C and Sheikh, Hamid R and Simoncelli, Eero P},
  journal={IEEE Transactions on Image Processing},
  volume={13},
  number={4},
  pages={600--612},
  year={2004},
  publisher={IEEE}
}

@article{lowe2004distinctive,
  title={Distinctive image features from scale-invariant keypoints},
  author={Lowe, David G},
  journal={International Journal of Computer Vision},
  volume={60},
  number={2},
  pages={91--110},
  year={2004},
  publisher={Springer}
}

@inproceedings{farneback2003two,
  title={Two-frame motion estimation based on polynomial expansion},
  author={Farneb{\"a}ck, Gunnar},
  booktitle={Scandinavian Conference on Image Analysis},
  pages={363--370},
  year={2003},
  organization={Springer}
}

@inproceedings{adam,
  title={Decoupled Weight Decay Regularization},
  author={Loshchilov, Ilya and Hutter, Frank},
  booktitle={International Conference on Learning Representations (ICLR)},
  year={2019},
  url={https://openreview.net/forum?id=Bkg6RiCqY7}
}

@inproceedings{resnet,
  title={Deep Residual Learning for Image Recognition},
  author={He, Kaiming and Zhang, Xiangyu and Ren, Shaoqing and Sun, Jian},
  booktitle={Proceedings of the IEEE Conference on Computer Vision and Pattern Recognition (CVPR)},
  pages={770--778},
  year={2016}
}

@article{imagenet,
  title={ImageNet: A large-scale hierarchical image database},
  author={Deng, Jia and Dong, Wei and Socher, Richard and Li, Li-Jia and Li, Kai and Fei-Fei, Li},
  journal={IEEE Conference on Computer Vision and Pattern Recognition (CVPR)},
  pages={248--255},
  year={2009}
}

@article{kolve2017ai2thor,
  title={AI2-THOR: An Interactive 3D Environment for Visual AI},
  author={Kolve, Eric and Mottaghi, Roozbeh and Gordon, Daniel and Zhu, Yuke and Gupta, Abhinav and Farhadi, Ali},
  journal={arXiv preprint arXiv:1712.05474},
  year={2017}
}

@inproceedings{depthanything,
  title     = {Depth Anything: Universal Monocular Depth Estimation},
  author    = {Yang, Shuai and Zhu, Yichao and Chen, Zhipeng and Chen, Wenguan},
  booktitle = {Proceedings of the IEEE/CVF Conference on Computer Vision and Pattern Recognition (CVPR)},
  year      = {2024}
}

@misc{so100_wiki,
  title        = {SO-100 Robotic Arm Quickstart Guide},
  author       = {{Seeed Studio}},
  howpublished = {\url{https://wiki.seeedstudio.com/lerobot_so100m/}},
  note         = {Accessed: 2025-02-14}
}
}


\end{document}